\documentclass{article}
\usepackage{amsmath,amsfonts,bm}

\def\eqref#1{equation~\ref{#1}}
\def\1{\bm{1}}

\DeclareMathAlphabet{\mathsfit}{\encodingdefault}{\sfdefault}{m}{sl}
\SetMathAlphabet{\mathsfit}{bold}{\encodingdefault}{\sfdefault}{bx}{n}

\usepackage{palatino}
\usepackage{amssymb}
\usepackage{amsmath}
\usepackage{amsthm}
\usepackage[a4paper]{geometry}
\usepackage[round]{natbib}
\usepackage[dvipsnames]{xcolor}
\usepackage[colorlinks,linktoc=true,
             citecolor= Cerulean!85!green!90!black,%Cerulean!75!green,
             linkcolor=YellowOrange,
             urlcolor=RubineRed!85!black,
             ]{hyperref}
\usepackage{graphicx}

\usepackage{hyperref}
\usepackage{url}
\usepackage{authblk}

\usepackage[utf8]{inputenc} % allow utf-8 input
\usepackage[T1]{fontenc}    % use 8-bit T1 fonts
\usepackage{hyperref}       % hyperlinks
\usepackage{url}            % simple URL typesetting
\usepackage{booktabs}       % professional-quality tables
\usepackage{amsfonts}       % blackboard math symbols
\usepackage{nicefrac}       % compact symbols for 1/2, etc.
\usepackage{microtype}      % microtypography
\usepackage{xcolor}         % colors

\usepackage{framed}
\usepackage{xcolor}
\usepackage{hyperref}
\usepackage{url}
\usepackage{listings}
\usepackage{enumitem}
\usepackage{microtype}
\usepackage{graphicx}
\usepackage{wrapfig}
\usepackage{caption}
\usepackage{subcaption}
\usepackage[utf8]{inputenc} % allow utf-8 input
\usepackage[T1]{fontenc}    % use 8-bit T1 fonts
\usepackage{hyperref}       % hyperlinks
\usepackage{url}            % simple URL typesetting
\usepackage{booktabs}       % professional-quality tables
\usepackage{amsfonts}       % blackboard math symbols
\usepackage{nicefrac}       % compact symbols for 1/2, etc.
\usepackage{microtype}      % microtypography
\usepackage{xcolor}         % colors

\usepackage{amsmath}
\usepackage{amssymb}
\usepackage{mathtools}
\usepackage{amsthm}
\usepackage{bm}
\usepackage{algorithm}
\usepackage{algorithmic}
\usepackage{color}
\usepackage{diagbox}
\usepackage{xcolor}
\usepackage{tikz}
\usepackage{tcolorbox}
\tcbuselibrary{breakable}
\theoremstyle{plain}

\theoremstyle{definition}

\theoremstyle{remark}

\usepackage[textsize=tiny]{todonotes}

\usepackage{multirow}



\title{Preference Tuning as Spectral Update Reorganization}

\makeatletter
\newcommand\email[2][]%
   {\newaffiltrue\let\AB@blk@and\AB@pand
      \if\relax#1\relax\def\AB@note{\AB@thenote}\else\def\AB@note{\relax}%
        \setcounter{Maxaffil}{0}\fi
      \begingroup
        \let\protect\@unexpandable@protect
        \def\thanks{\protect\thanks}\def\footnote{\protect\footnotetext{\textdagger\ #1}}%
        \@temptokena=\expandafter{\AB@authors}%
        {\def\\{\protect\\\protect\Affilfont}\xdef\AB@temp{#2}}%
         \xdef\AB@authors{\the\@temptokena\AB@las\AB@au@str
         \protect\\[\affilsep]\protect\Affilfont\AB@temp}%
         \gdef\AB@las{}\gdef\AB@au@str{}%
        {\def\\{, \ignorespaces}\xdef\AB@temp{#2}}%
        \@temptokena=\expandafter{\AB@affillist}%
        \xdef\AB@affillist{\the\@temptokena \AB@affilsep
          \AB@affilnote{}\protect\Affilfont\AB@temp}%
      \endgroup
       \let\AB@affilsep\AB@affilsepx
}
\makeatother

\author[1\thanks{Contact: <pzhangao@cse.ust.hk>}]{Peiyan Zhang}
\author[2]{Haibo Jin}
\author[3]{Liying Kang}
\author[2]{Haohan Wang}
\affil[1]{Hong Kong University of Science and Technology}
\affil[2]{University of Illinois at Urbana-Champaign}
\affil[3]{Hong Kong Polytechnic University}
\date{}

\begin{document}
\maketitle

\begin{abstract}

Preference-based post-training is usually understood through endpoint behavior, yet the learned update that produces this behavior remains largely opaque. We study RLHF and related preference optimization through the spectral structure of their induced parameter updates. By decomposing effective LoRA updates and reloading their spectral components as plug-in modules, we turn preference-induced updates into objects that can be isolated, recomposed, and directly intervened on.
Across model families, optimization algorithms, and supervision regimes, these updates consistently develop a spectral head--tail organization. A compact head emerges early and carries the dominant endpoint shift, while a heterogeneous residual tail remains. The split is functional rather than merely descriptive. Plug-in intervention shows that the head accounts for the visible behavioral departure from the base model, while the tail is weak in isolation. Cross-run recomposition further shows that mixed adapters follow the source of the head, indicating that the head carries run-level solver bias.
This endpoint dominance does not imply learning sufficiency. Head-only learning is non-vacuous but fails to recover the full solution, especially on out-of-distribution behavior. Tail-only learning yields little visible gain, yet the full solution is not recovered without the tail. These findings recast preference post-training as structured update reorganization rather than a monolithic behavioral correction, and suggest that alignment gain and coverage loss are tied to how the learned update itself is organized.

% Reinforcement learning from human feedback (RLHF) can substantially improve the behavior of large language models (LLMs), yet the internal structure of the updates it induces remains poorly understood. We study RLHF through the spectral structure of the parameter update between an RLHF-tuned model and its pre-RLHF checkpoint. Across model families, RL algorithms, and random seeds, we find that RLHF updates consistently develop a stage-like spectral separation over training: a coherent low-rank component emerges early and supports broad, prompt-consistent behavioral shifts, while a later high-rank tail remains more heterogeneous and is associated with conditional specialization and capability breadth. We further show that this split is behaviorally meaningful rather than merely descriptive. Under source-controlled intervention, the low component carries the dominant and transferable behavioral payload, whereas the high component is weak in isolation. Yet low dominance at the final checkpoint does not imply training-time sufficiency: restricting learning to the low-spectral subspace fails to recover the full solution, especially on out-of-distribution behavior. Taken together, these results suggest that RLHF is better understood not as a single uniform update, but as a structured reorganization whose components play different and sometimes competing roles. This perspective turns the recurring tension between alignment gain and coverage loss into an internal property of the learned update itself.

\end{abstract}

\section{Introduction}
\label{sec:intro}

Preference-based post-training has become a standard mechanism for shaping the behavior of modern large language models (LLMs)~\citep{christiano2017deep,ouyang2022training,schulman2017proximal,rafailov2023direct}. It is typically judged by endpoint behavior: whether the tuned model follows instructions more reliably, produces preferred responses, or improves on reasoning-oriented tasks~\citep{jaech2024openai,guo2025deepseek,team2025kimi}. Yet this endpoint view leaves the learned update itself largely opaque. Post-training does not merely change model outputs; it induces a parameter update whose internal organization may determine which behaviors are strengthened, preserved, or lost. This distinction matters because preference tuning often improves broadly shared alignment while weakening more conditional, specialized, or long-tail capabilities~\citep{liu2025reinforcement}. Explaining this tension, therefore, requires looking beyond whether the final model improves, and asking how the induced update is structured inside the model.

Prior work frames this issue as a debate about endpoint behavior: whether preference tuning merely reweights capabilities already present in the base model, or whether it creates new reasoning and decision patterns~\citep{mukherjee2025reinforcement,yue2025does,zhang2025reinforcement,chen2025retaining,wang2025octothinker}. This framing misses a more basic structural question. A preference-tuned model differs from its initialization by a learned update, yet that update is usually treated as a monolithic consequence of optimization. Consequently, it remains unclear whether alignment gains, solver biases, and coverage losses arise from the same update directions or from separable components with different roles.

We therefore make the learned update the unit of analysis. For each preference-tuned checkpoint, we treat its difference from the pre-tuning initialization as an additive effective update, instantiated in our main experiments by LoRA adapters~\citep{hu2022lora}. We decompose each module-level update with singular value decomposition (SVD)~\citep{wall2003singular}, yielding a leading \emph{spectral head} and a complementary \emph{residual tail}. These components are not only descriptive: they can be converted back into plug-in adapters, enabling controlled deletion, isolation, cross-run recomposition, and training-time restriction. This turns preference-induced updates from opaque optimization endpoints into manipulable experimental objects.

This update-centered view leads to a different picture of preference post-training. Rather than acting as a single monolithic behavioral correction, preference tuning reorganizes the learned update into structurally distinct components. A dominant spectral head consolidates the main behavioral shift, while a heterogeneous residual tail remains weak in isolation but is tied to the recovery of broader and more conditional behavior. Under this view, the tension between alignment gain and coverage loss is not merely an external side effect of post-training, but a property of how the update itself is organized.

We substantiate this view through three findings.

\begin{itemize}[leftmargin=*,noitemsep,topsep=0pt]
    \item \textbf{Preference-induced updates develop a stable spectral head--tail structure.}
    Across model families, optimization algorithms, and supervision regimes, the update does not remain diffuse or collapse into a purely low-rank object. A coherent spectral head emerges early, while a heterogeneous residual tail remains throughout training.

    \item \textbf{The spectral split is functional, not merely descriptive.}
    Source-controlled plug-in intervention shows that the spectral head and residual tail induce different endpoint behaviors when isolated. The residual tail remains weak in isolation, whereas the spectral head carries the dominant but run-dependent behavioral effect.

    \item \textbf{Functional separation has mechanistic signatures across recomposition, training, and supervision.}
    Cross-run recomposition shows that mixed adapters follow the source of the spectral head, indicating that the head carries run-level solver bias. Training-time projection shows that head-only learning is non-vacuous but fails to recover the full solution, especially on out-of-distribution behavior. Supervision corruption further suggests that coherent prompt--preference alignment helps determine whether coherent update structure can form.
\end{itemize}

Together, these findings frame preference post-training as structured update reorganization rather than a single behavioral shift. The spectral head explains the dominant endpoint behavior, but the residual tail and the supervision conditions under which the structure forms are essential to understanding breadth, learnability, and coverage loss.
% \hwc{we probably need a related work section.}

\section{Related Work}
\label{sec:related_work}

\textbf{Preference post-training and behavioral accounts.}
Preference-based post-training has become a central mechanism for shaping the behavior of large language models. Early work studied reward learning from human preference comparisons~\citep{christiano2017deep}. RLHF later became a standard recipe for instruction alignment~\citep{ouyang2022training,bai2022training}. Direct preference optimization simplified this pipeline by optimizing directly from preference pairs without an explicit reward-model reinforcement-learning loop~\citep{rafailov2023direct}. Recent reasoning-oriented post-training extends reward-based optimization to mathematical, coding, and complex reasoning tasks. Representative examples include GRPO-style training in DeepSeekMath~\citep{shao2024deepseekmath} and later reasoning models trained with large-scale reinforcement learning~\citep{guo2025deepseek}. These works define the class of post-training processes we study, but they are primarily evaluated through endpoint behavior, such as instruction following, preferred response generation, and reasoning accuracy. We do not propose a new preference objective. We analyze the parameter update induced by such objectives and ask how this update is organized inside the model.

\noindent\textbf{What RLHF changes in language models.}
Recent studies have begun to question what endpoint improvement reflects. One line of work asks whether RLVR truly expands the base model's reasoning capacity or mainly improves the sampling efficiency of high-reward paths already present in the base distribution~\citep{yue2025does}. Other work analyzes whether the reasoning boundary shrinks, expands, or changes in a stage-dependent manner during training~\citep{yao2025debate}. Capability-boundary collapse in RLVR has also motivated hybrid-policy and exploration-based methods that aim to counteract such narrowing~\citep{dong2025rl}.

Another line localizes RL-induced changes at finer granularity. Some studies analyze how RL reshapes reasoning patterns and token-level dynamics~\citep{chen2025reshaping}. Others show that high-entropy or high-significance decision tokens play a disproportionate role in RL-based reasoning~\citep{wang2025beyond,liu2026tokensmatterefficientllm}. Trajectory-level information peaks have also been used to identify key thinking tokens in reasoning traces~\citep{qian2025demystifying}. At the parameter level, recent work finds that RL fine-tuning can modify sparse subnetworks rather than the full model~\citep{mukherjee2025reinforcement,balashov2025reinforcement}, and that random sparse subnetworks may suffice for effective RLVR~\citep{adewuyi2026multiple}. Work on RLVR update directions further argues that update magnitude or sparsity alone is insufficient for understanding RL-induced change~\citep{huang2026direction}. Other studies examine base-model compatibility, mid-training conditions, and scaling behavior in RL post-training~\citep{wang2025octothinker,tan2025scaling}. Post-training trade-offs such as forgetting and retention have also been studied, with evidence that online RL can preserve prior capabilities differently from supervised fine-tuning~\citep{chen2025retaining,shenfeld2025rl}.

These studies suggest that RL effects are not uniform endpoint-score improvements. They appear across reasoning trajectories, token-level decision points, parameter footprints, update directions, and capability coverage. Most analyses, however, still operate at the level of behavioral trajectories, token dynamics, sparse subnetworks, or coarse update effects. We instead study whether a preference-induced update contains separable spectral components with distinct endpoint and learning roles.

\noindent\textbf{Parameter-space views of model behavior.}
A complementary line of work treats the difference between a fine-tuned model and its pretrained initialization as a manipulable object in parameter space. Task arithmetic defines such differences as task vectors and composes them to steer model behavior~\citep{ilharco2022editing}. Model soups average fine-tuned weights to improve accuracy without increasing inference cost~\citep{wortsman2022model}. Later merging methods study how to reduce interference when combining task-specific models~\citep{yadav2023ties}. Recent work further develops task-vector bases for efficient model editing~\citep{zeng2025efficient}, layer-aware task arithmetic for disentangling task-specific knowledge from instruction-following behavior~\citep{chen2025layer}, and task singular vectors for reducing task interference in model merging~\citep{gargiulo2025task}.

We share the view that fine-tuning differences are meaningful objects for analysis and intervention, but prior work typically manipulates the update as a whole for model editing or merging. Closest in spirit, task singular vector methods also use SVD to analyze task matrices~\citep{gargiulo2025task}, but their goal is to reduce interference in model merging. Our spectral components are used instead as diagnostic and interventional objects for functional separation within a preference-induced update, allowing us to distinguish endpoint salience from learning sufficiency.

\section{Preliminaries}
% \vspace{-5pt}
\label{sec:preliminaries}

Endpoint behavior collapses the effect of preference tuning into a final model response. This view is useful for evaluation, but it cannot reveal whether the same learned directions control the visible behavioral shift, support optimization, or preserve more conditional capabilities. We therefore use the additive update from the pre-tuning model to the tuned checkpoint as the basic object of analysis. Once written as a collection of module-level patches, this update can be decomposed, removed, substituted, and recomposed without changing the base model.

\subsection{Effective Updates}
\label{sec:effective_updates}

Module-level linear projections provide a natural unit for this analysis, since the learned change to each projection can be expressed as an additive patch on the corresponding base weight. Let \(W_{\ell,j}^{(0)}\) denote the pre-tuning weight of module \(j\) in Transformer block \(\ell\), and let \(W_{\ell,j}^{(t)}\) denote the corresponding weight at checkpoint \(t\). The effective update \(\Delta W_{\ell,j}^{(t)}\) is defined by
\[
W_{\ell,j}^{(t)}
=
W_{\ell,j}^{(0)}
+
\Delta W_{\ell,j}^{(t)} .
\]
For dense finetuning, \(\Delta W_{\ell,j}^{(t)} = W_{\ell,j}^{(t)} - W_{\ell,j}^{(0)}\). In our main LoRA setting~\citep{hu2022lora}, the backbone is frozen and the effective update is given by the scaled adapter product
\[
\Delta W_{\ell,j}^{(t)}
=
s_{\ell,j}^{(t)}
B_{\ell,j}^{(t)}
A_{\ell,j}^{(t)},
\]
where \(A_{\ell,j}^{(t)}\) and \(B_{\ell,j}^{(t)}\) are the LoRA factors and \(s_{\ell,j}^{(t)}\) absorbs implementation-specific scaling.

The checkpoint-level update is the collection of these module-level patches over the targeted attention and MLP projections. Full module lists, LoRA dimensions, and scaling conventions are given in Appendix~\ref{app:update_details}. Because the update is additive, component deletion, isolation, cross-run substitution, and recomposition can all be implemented as operations on \(\Delta W\). When no ambiguity arises, we omit the checkpoint superscript \(t\).

\subsection{Spectral Partition of Updates}
\label{sec:spectral_head_tail}

A useful partition must separate dominant update directions while preserving exact recomposability. For each module-level update \(\Delta W_{\ell,j}\), we compute
\[
\Delta W_{\ell,j}
=
U_{\ell,j}\Sigma_{\ell,j}V_{\ell,j}^{\top},
\]
with singular values sorted in decreasing order.

Given a split rank \(r\), the spectral head is the leading truncated component
\[
\Delta W_{\ell,j}^{\mathrm{head}}(r)
=
U_{\ell,j}^{(:,1:r)}
\Sigma_{\ell,j}^{(1:r,1:r)}
\bigl(V_{\ell,j}^{(:,1:r)}\bigr)^{\top},
\]
and the residual tail is its complement
\[
\Delta W_{\ell,j}^{\mathrm{tail}}(r)
=
\Delta W_{\ell,j}
-
\Delta W_{\ell,j}^{\mathrm{head}}(r).
\]
Thus,
\[
\Delta W_{\ell,j}
=
\Delta W_{\ell,j}^{\mathrm{head}}(r)
+
\Delta W_{\ell,j}^{\mathrm{tail}}(r),
\]
which enables exact deletion, isolation, and recomposition at the update level.

The names head and tail are purely spectral at definition time. We do not assume that they correspond to fixed behavioral categories. Their roles are tested empirically through plug-in intervention, cross-run recomposition, and training-time projection.

\subsection{Tracking Spectral Concentration}
\label{sec:spectral_statistics}

To study whether preference tuning forms a concentrated update structure over training, we track the singular-value spectra of each \(\Delta W_{\ell,j}^{(t)}\) across checkpoints and summarize concentration with entropy effective rank. For singular values \(\{\sigma_i\}\), define
\[
p_i
=
\frac{\sigma_i}{\sum_k \sigma_k},
\qquad
r_{\mathrm{eff}}
=
\exp\left(
-\sum_i p_i \log(p_i+\varepsilon)
\right),
\]
where \(\varepsilon\) is a small numerical constant.

Smaller effective rank indicates that update mass is concentrated in fewer dominant directions, while larger effective rank indicates a flatter and more dispersed spectrum. Unless otherwise stated, checkpoint-level summaries report the median \(r_{\mathrm{eff}}\) over targeted modules. Numerical safeguards and additional rank statistics are provided in Appendix~\ref{app:update_details}.

\section{Spectral Separation}
\label{sec:spectral_separation}
% \vspace{-10pt}

A head--tail partition should not be assumed to be meaningful from the decomposition alone. The learned update could remain broadly diffuse, with no stable leading component, or collapse into a few dominant directions, leaving little residual structure to analyze. Preference-induced updates follow neither pattern. During training, update mass concentrates into a compact spectral head, while a non-negligible residual tail remains.

We first examine a controlled reference trajectory to trace how this structure forms over training, then test whether the same pattern persists across model families and supervision regimes.

\subsection{Formation in a Reference Trajectory}
\label{sec:spectral_reference}

We start from a controlled reference trajectory, \textbf{Qwen2.5-0.5B} under synthetic preference supervision, to isolate the formation of spectral structure before testing its generality. The backbone is frozen and preference tuning is performed with LoRA adapters of rank \(r_0=64\). For each checkpoint, we track the module-level effective updates defined in Section~\ref{sec:effective_updates} through two complementary views: singular-value spectra, which show the shape of each update operator, and entropy effective rank, which summarizes spectral concentration. Full training details, target modules, and checkpoint schedules are given in Appendix~\ref{app:spectral_full}.

\begin{figure}[t]
  \centering
  \begin{subfigure}[t]{0.48\linewidth}
    \centering
        % \vspace{-6pt}
    \includegraphics[width=\linewidth]{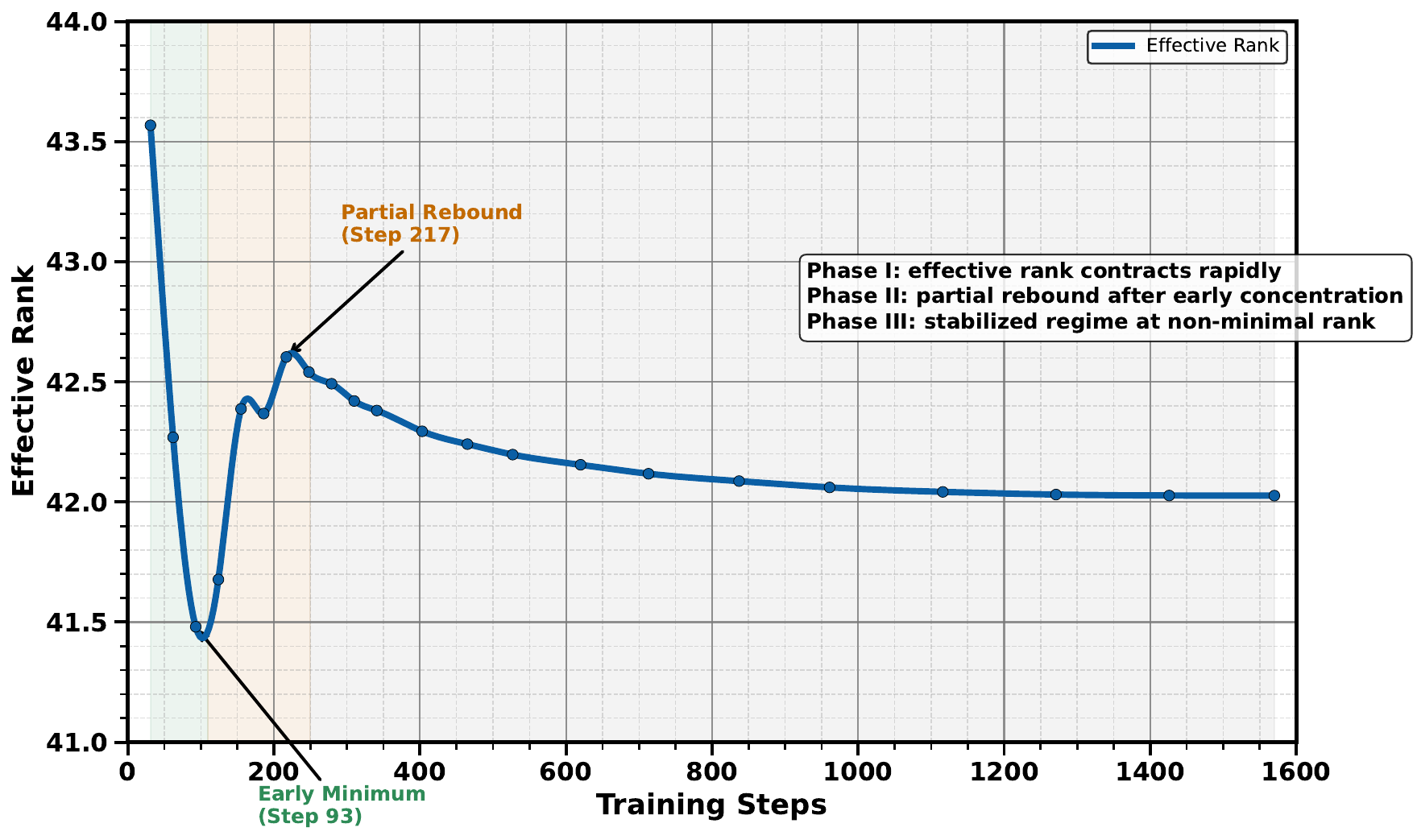}
    \caption{Effective rank}
  \end{subfigure}
  \hfill
  \begin{subfigure}[t]{0.48\linewidth}
    \centering
        % \vspace{-6pt}
    \includegraphics[width=\linewidth]{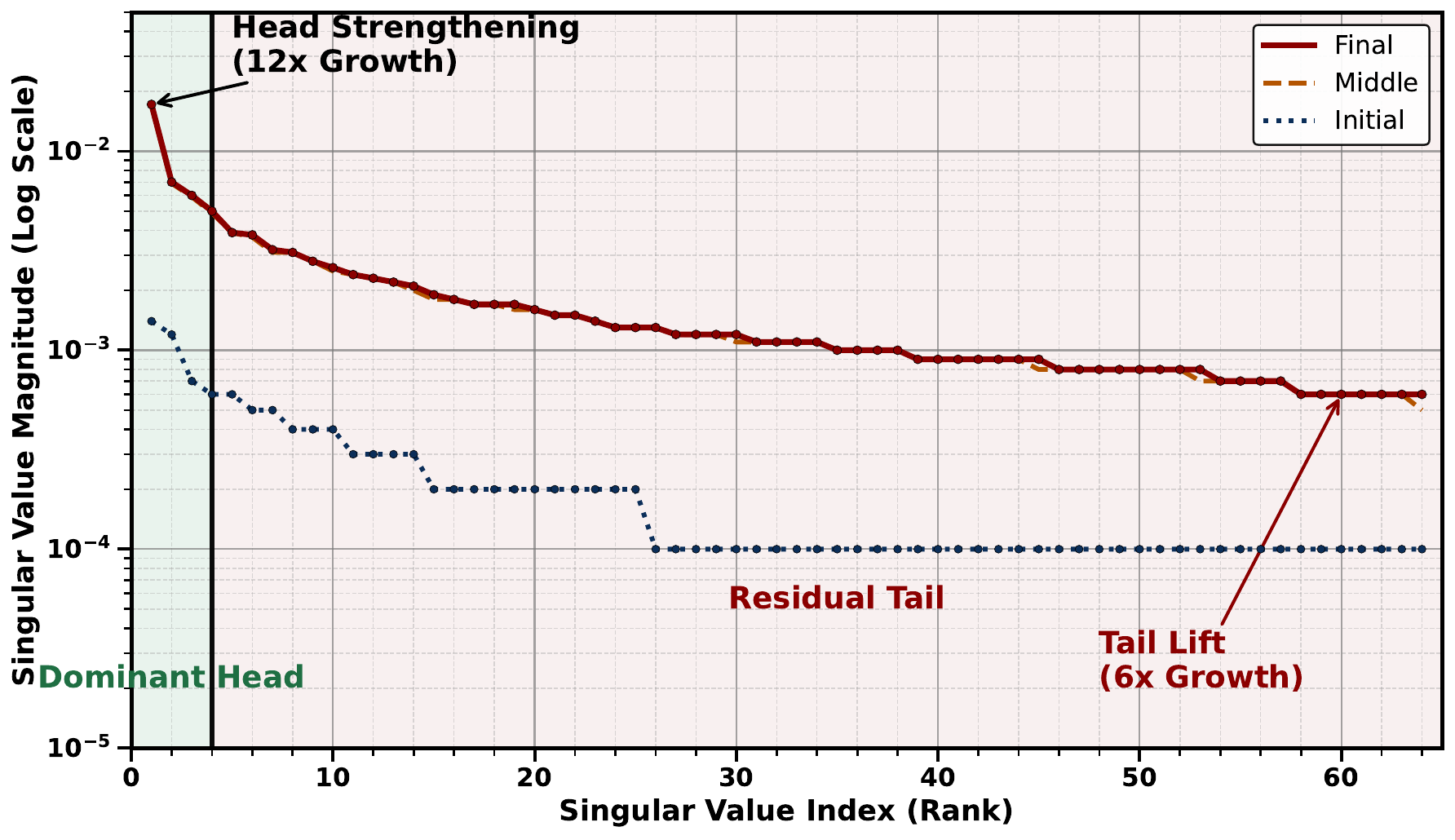}
    \caption{Singular spectra}
  \end{subfigure}
    % \vspace{-6pt}
  \caption{
  Formation of spectral head--tail organization in the reference trajectory.
  The effective rank contracts early, partially rebounds, and then stabilizes, while the singular spectra show a growing dominant head and a persistent residual tail.
  }
  % \vspace{-15pt}
  \label{fig:spectral_reference}
\end{figure}

Figure~\ref{fig:spectral_reference} shows that spectral organization forms early but does not continue toward rank collapse. The effective rank first decreases, indicating that update mass concentrates into fewer leading directions. It then partially rebounds and stabilizes, showing that preference tuning does not keep compressing the update into an increasingly smaller subspace.

The singular spectra explain this non-monotonic trajectory. Early in training, the spectrum is relatively flat and weakly structured. As training proceeds, the leading singular values separate from the rest of the spectrum and form a compact head. At the same time, the remaining singular directions do not disappear. The residual tail stays visible, accounting for the rebound and stabilization in effective rank. Preference tuning therefore reorganizes the update into a head--tail spectrum rather than leaving it diffuse or reducing it to a purely low-rank object.

\begin{figure*}[h]
\centering

\setlength{\tabcolsep}{2.5pt}
\renewcommand{\arraystretch}{1.0}

\begin{tabular}{@{}c c c c c@{}}
&
\shortstack[c]{\footnotesize DPO \\ \footnotesize Synthetic}
&
\shortstack[c]{\footnotesize DPO \\ \footnotesize Benchmark-derived}
&
\shortstack[c]{\footnotesize GRPO \\ \footnotesize Synthetic}
&
\shortstack[c]{\footnotesize GRPO \\ \footnotesize Benchmark-derived}
\\[-0.2em]

\rotatebox{90}{\hspace{0.2em}\footnotesize Qwen-1.7B} &
\begin{subfigure}[t]{0.215\textwidth}
    \includegraphics[width=\linewidth]{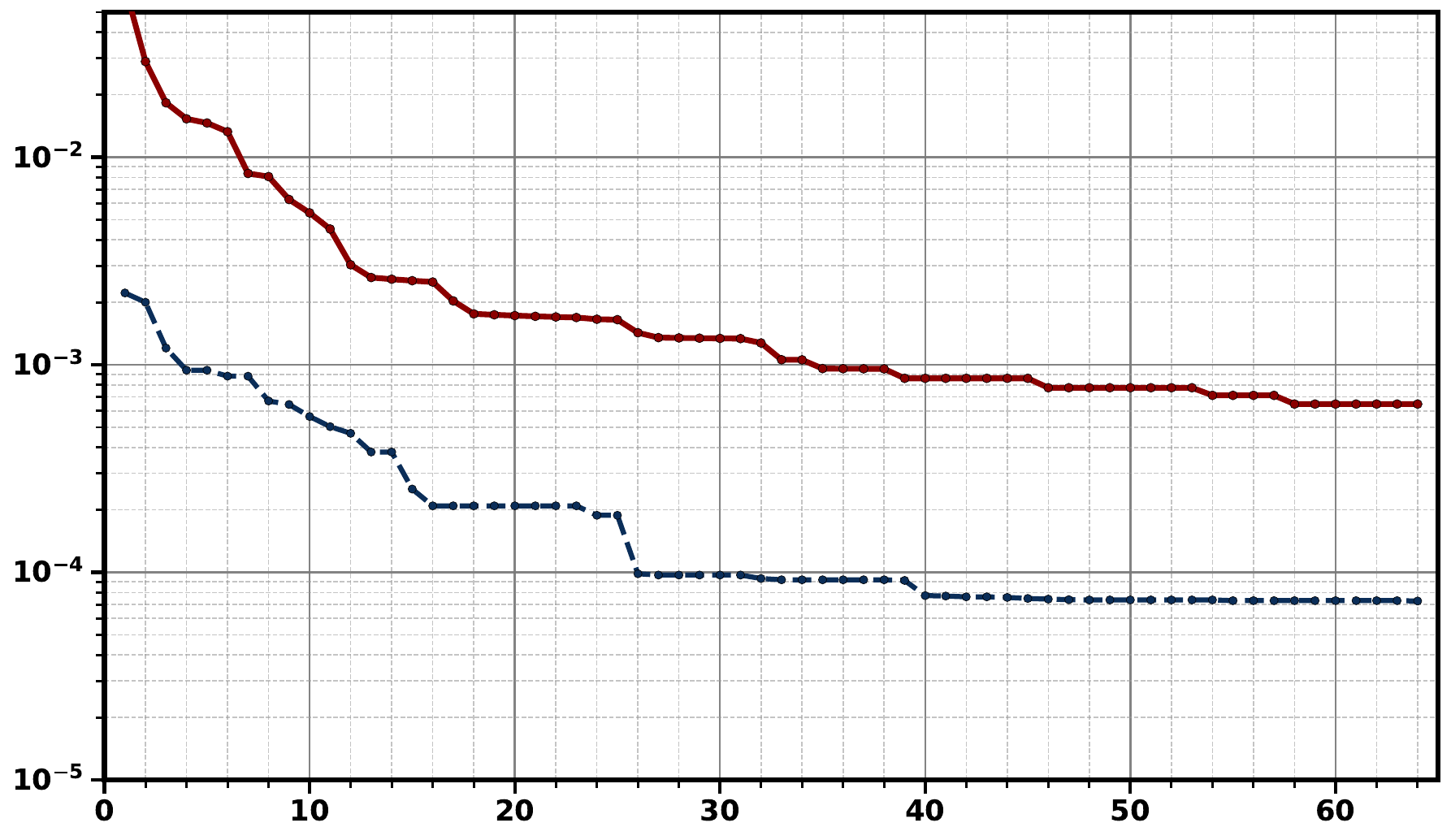}
\end{subfigure} &
\begin{subfigure}[t]{0.215\textwidth}
    \includegraphics[width=\linewidth]{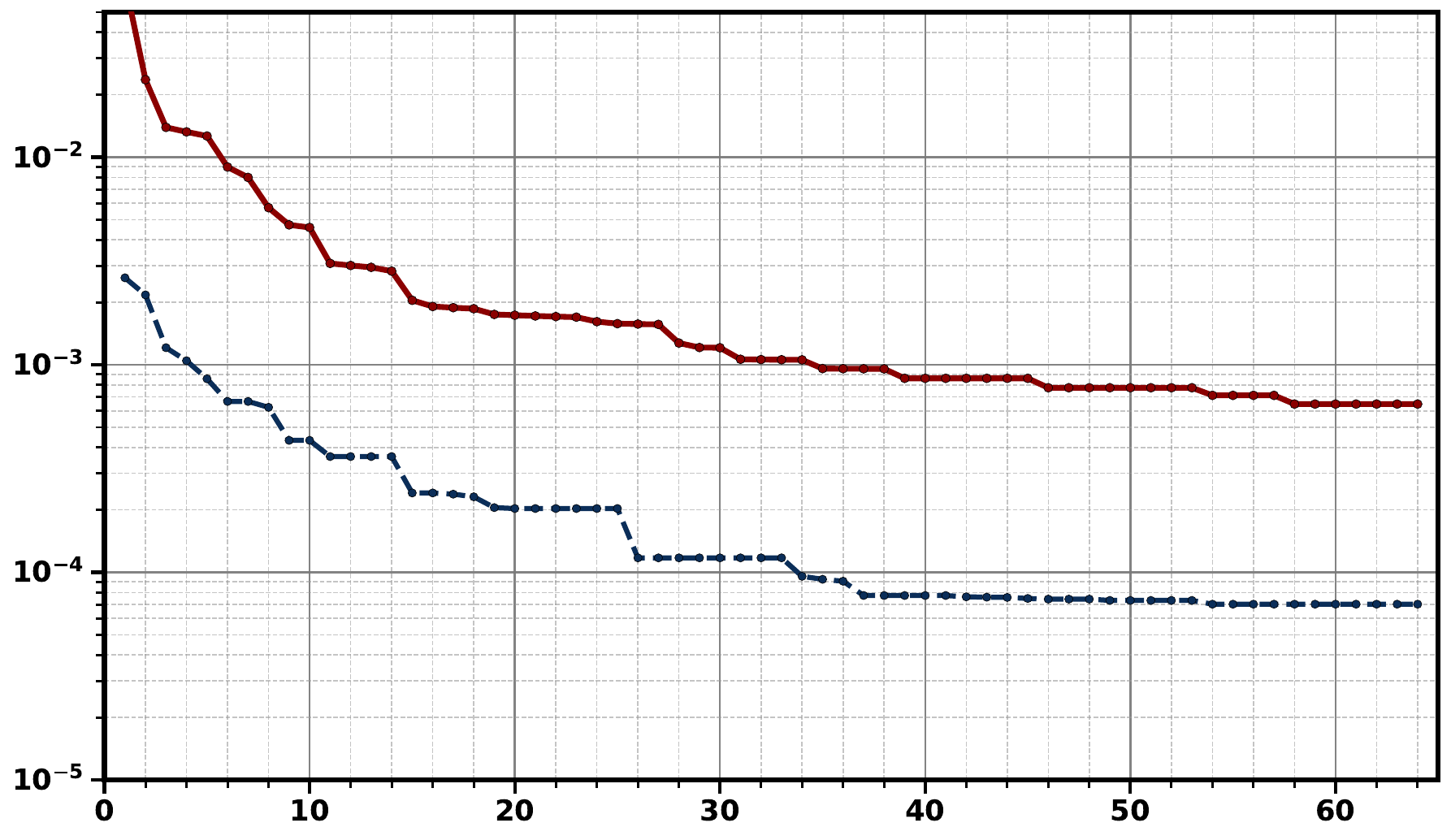}
\end{subfigure} &
\begin{subfigure}[t]{0.215\textwidth}
    \includegraphics[width=\linewidth]{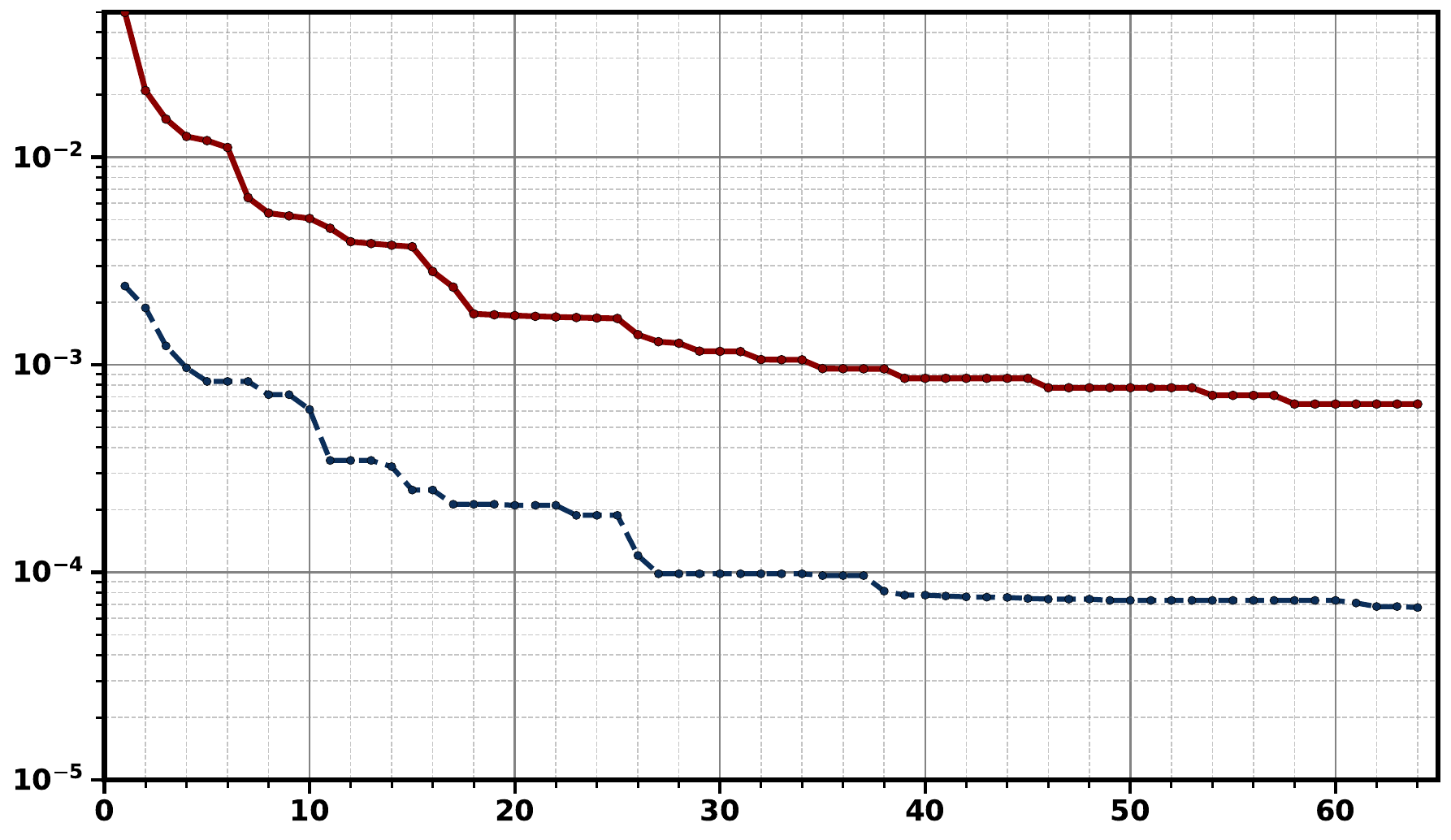}
\end{subfigure} &
\begin{subfigure}[t]{0.215\textwidth}
    \includegraphics[width=\linewidth]{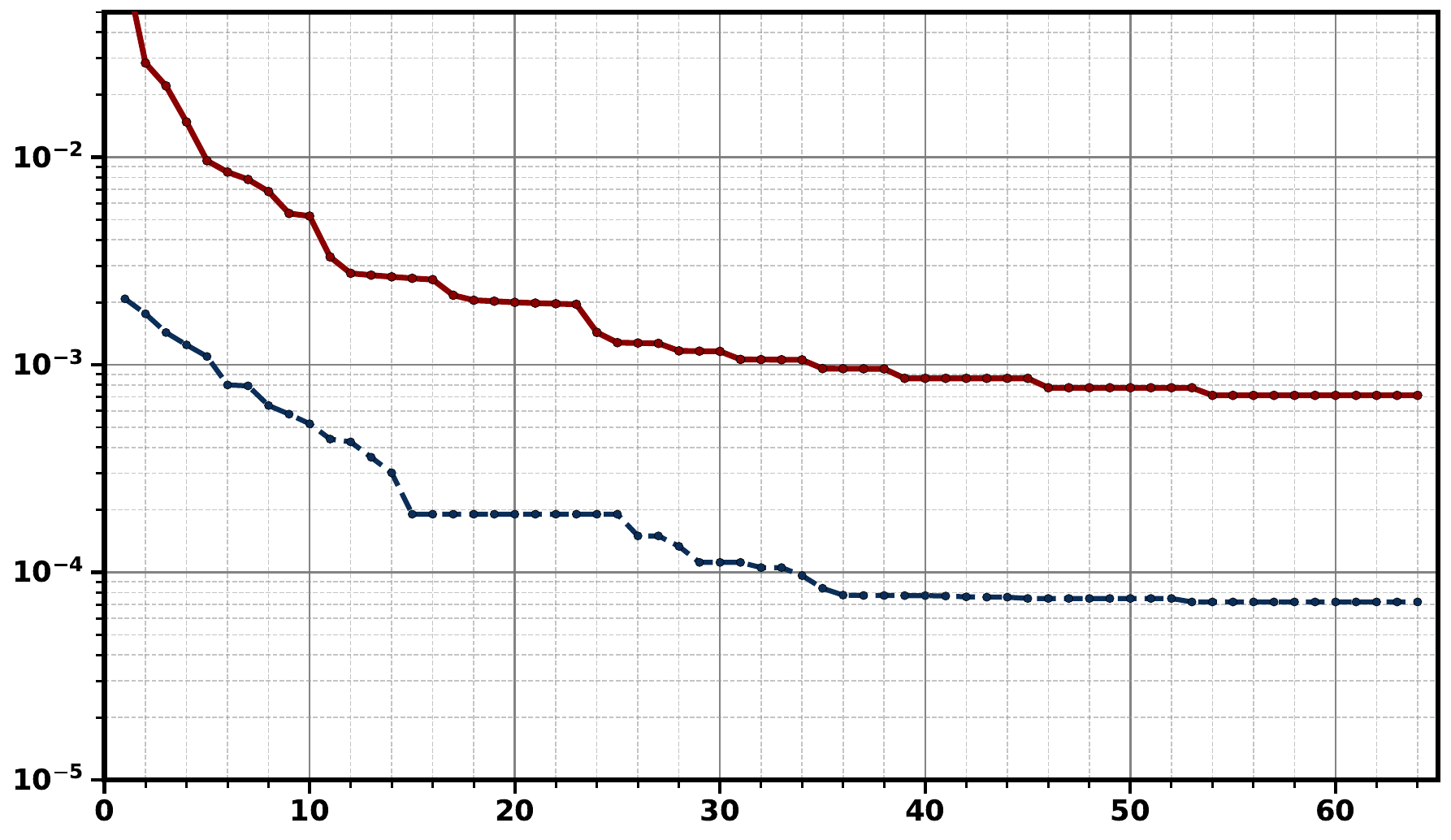}
\end{subfigure}
\\[-0.1em]

\rotatebox{90}{\hspace{0.2em}\footnotesize Qwen-8B} &
\begin{subfigure}[t]{0.215\textwidth}
    \includegraphics[width=\linewidth]{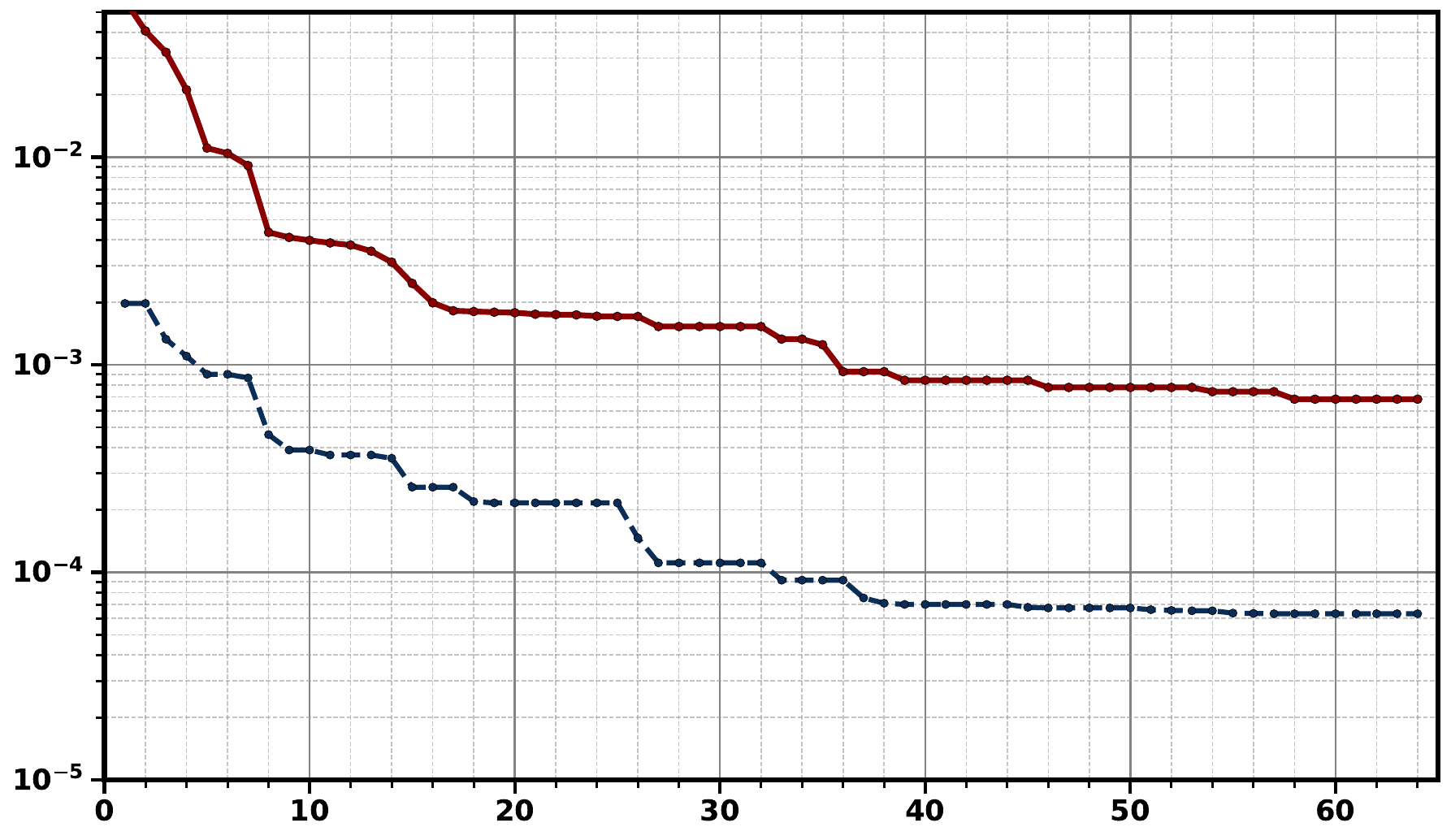}
\end{subfigure} &
\begin{subfigure}[t]{0.215\textwidth}
    \includegraphics[width=\linewidth]{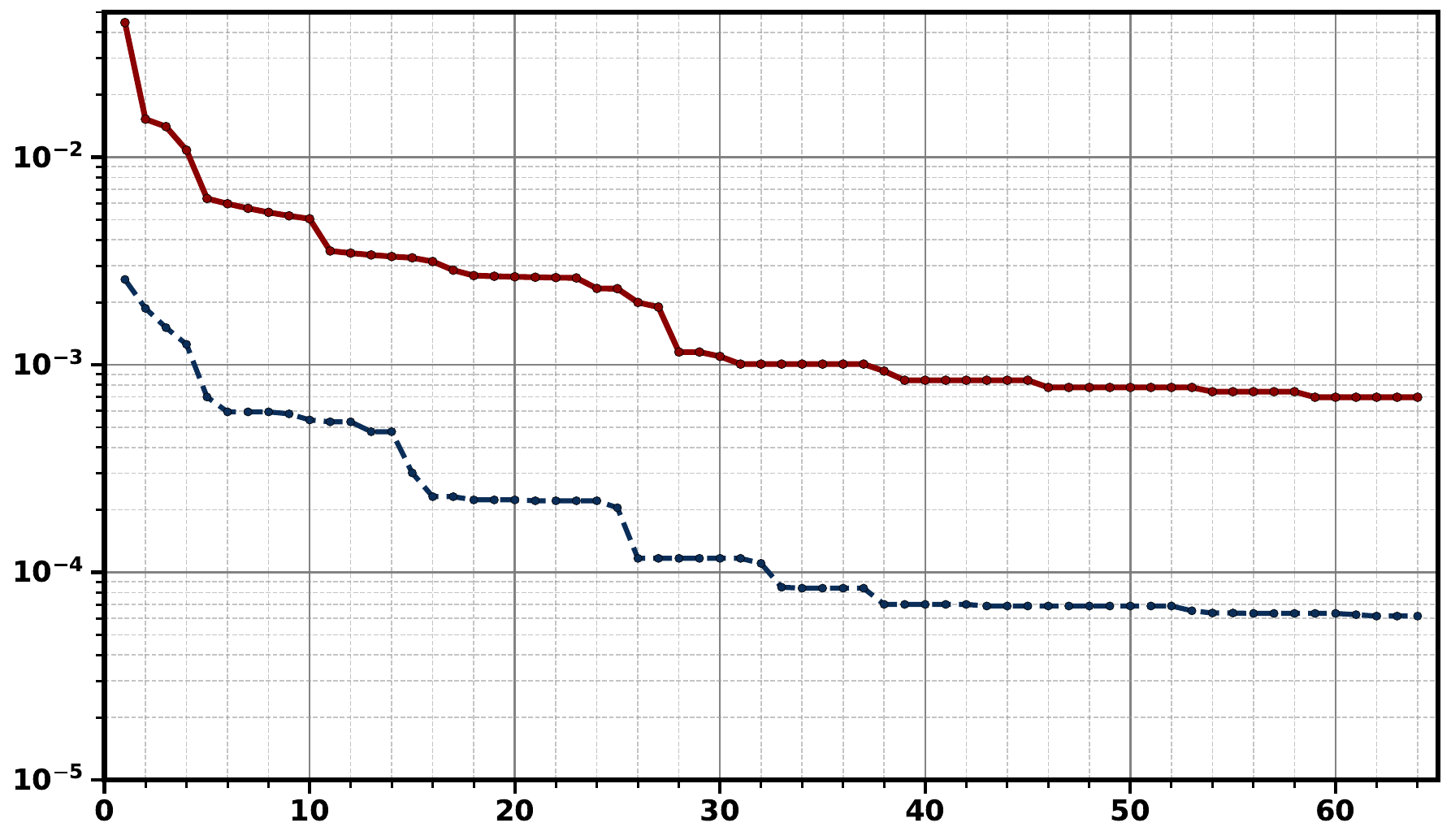}
\end{subfigure} &
\begin{subfigure}[t]{0.215\textwidth}
    \includegraphics[width=\linewidth]{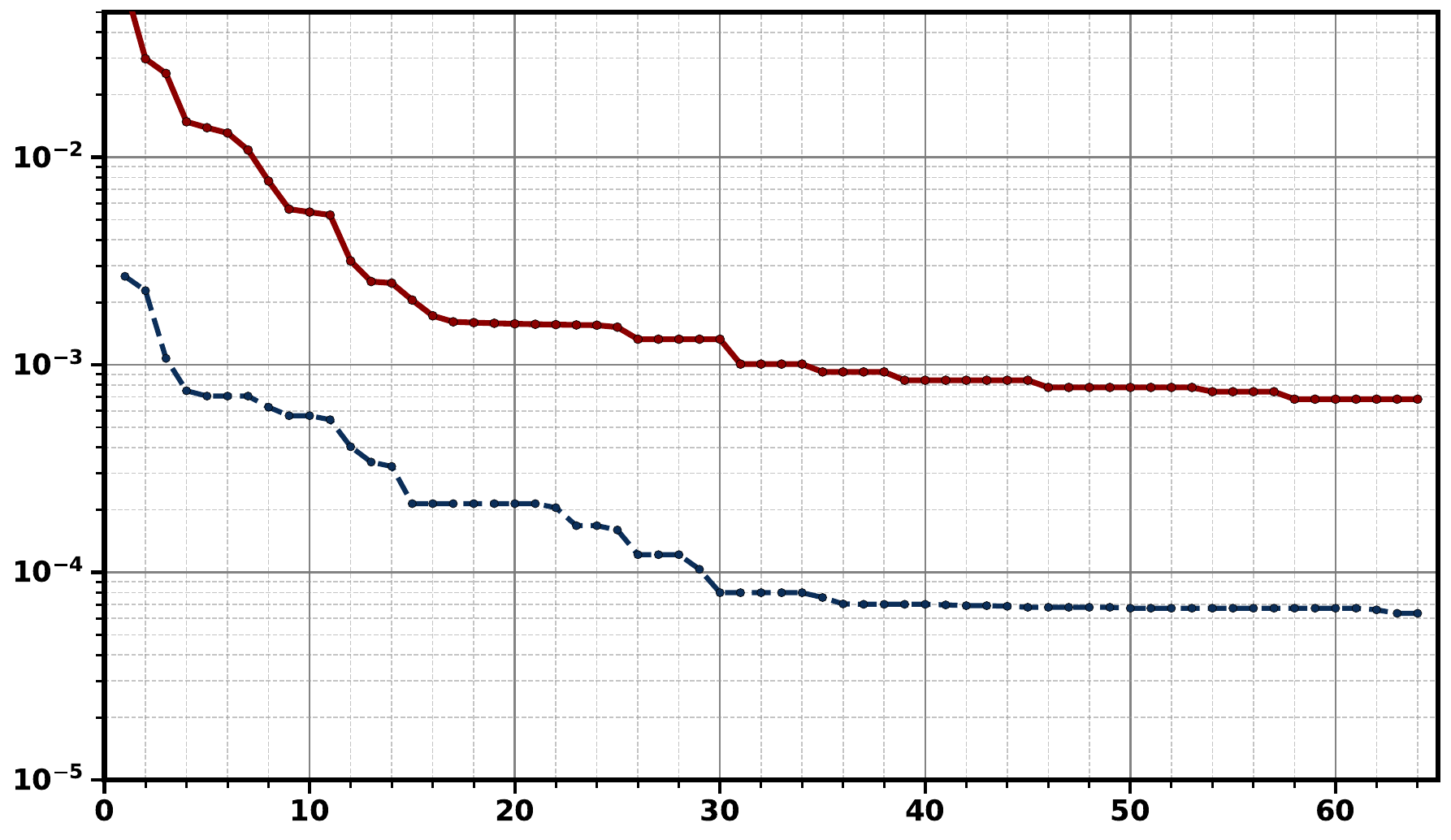}
\end{subfigure} &
\begin{subfigure}[t]{0.215\textwidth}
    \includegraphics[width=\linewidth]{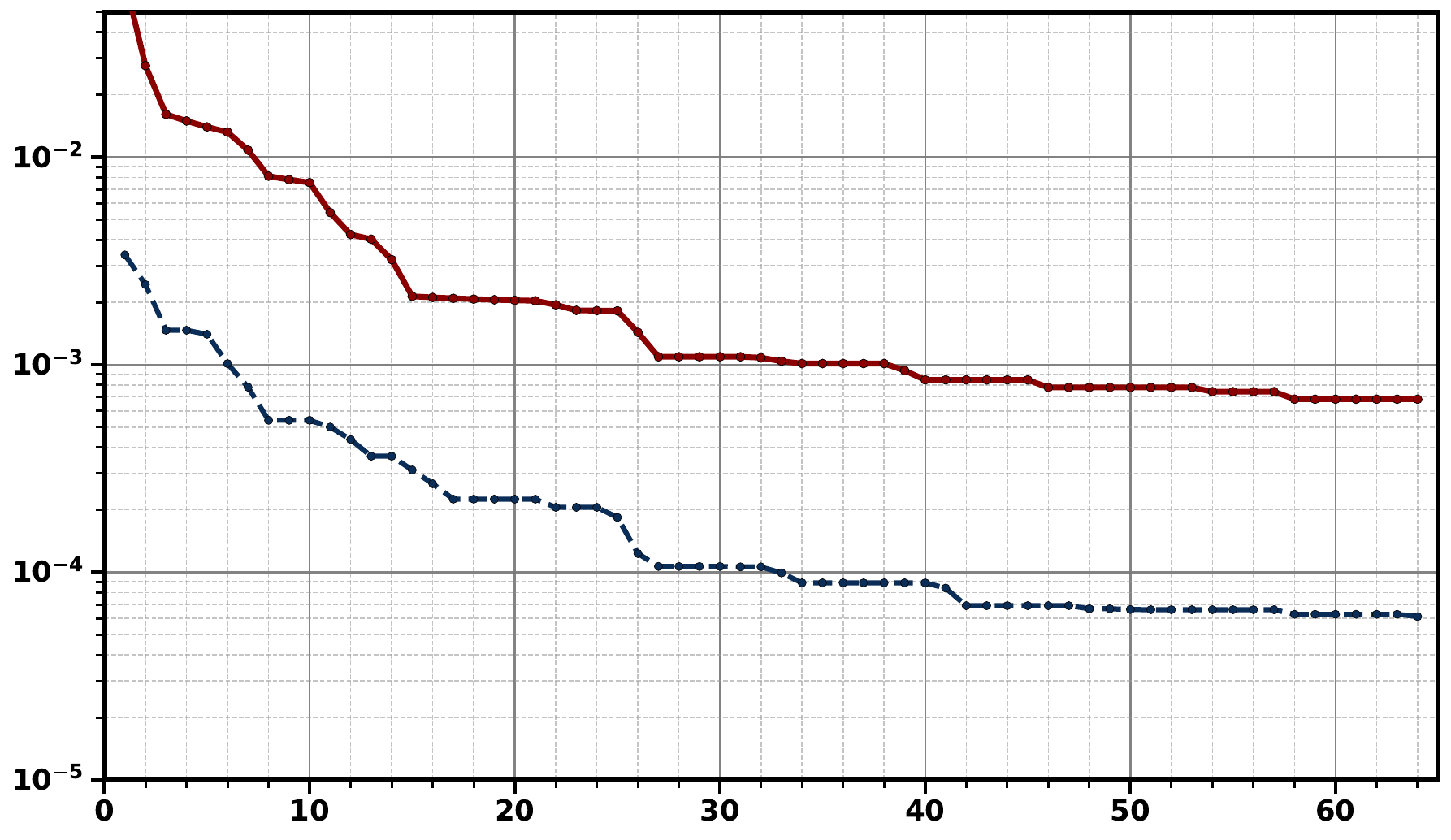}
\end{subfigure}
\\[-0.1em]

\rotatebox{90}{\hspace{0.2em}\footnotesize Llama-3.1-8B} &
\begin{subfigure}[t]{0.215\textwidth}
    \includegraphics[width=\linewidth]{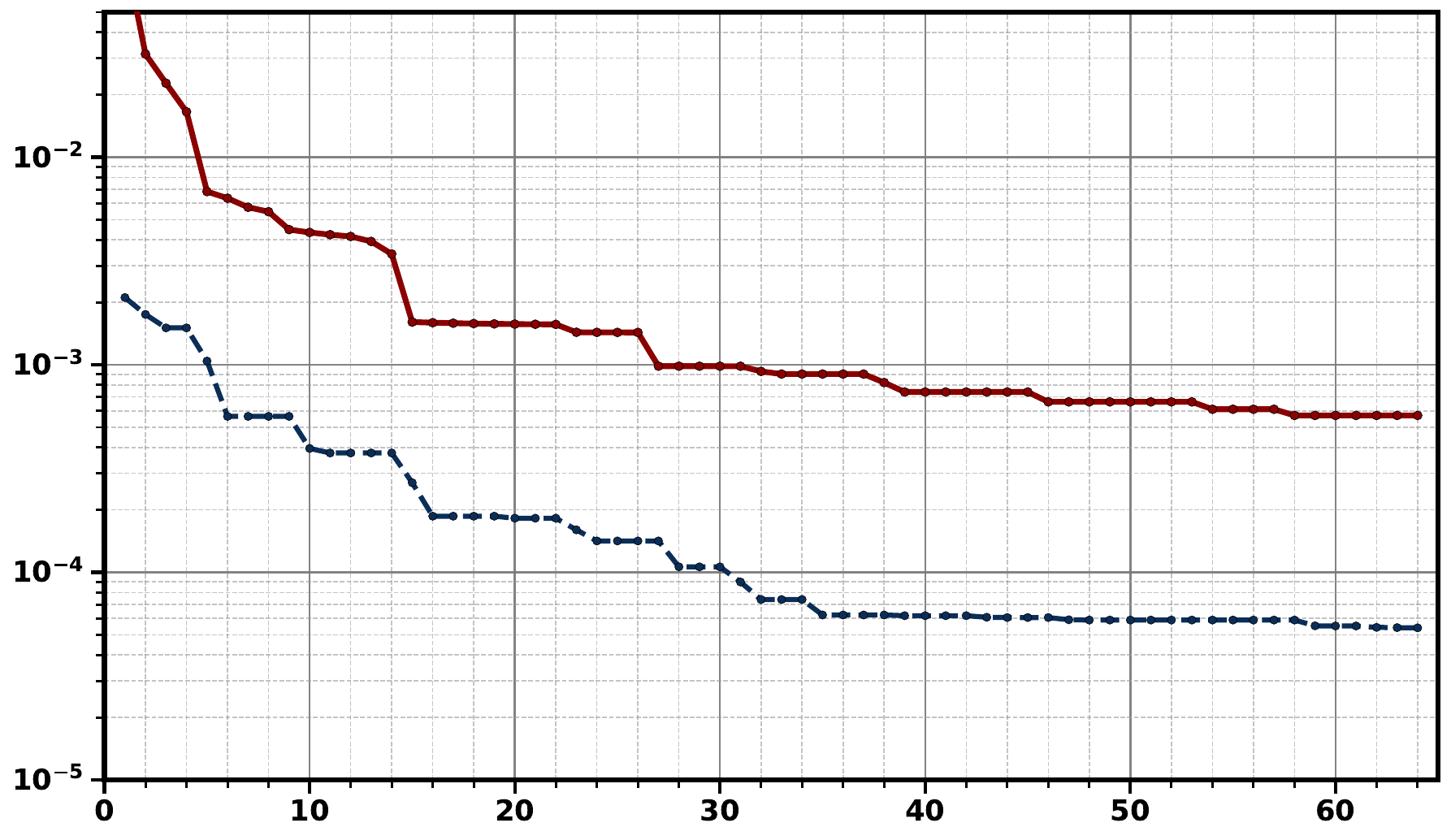}
\end{subfigure} &
\begin{subfigure}[t]{0.215\textwidth}
    \includegraphics[width=\linewidth]{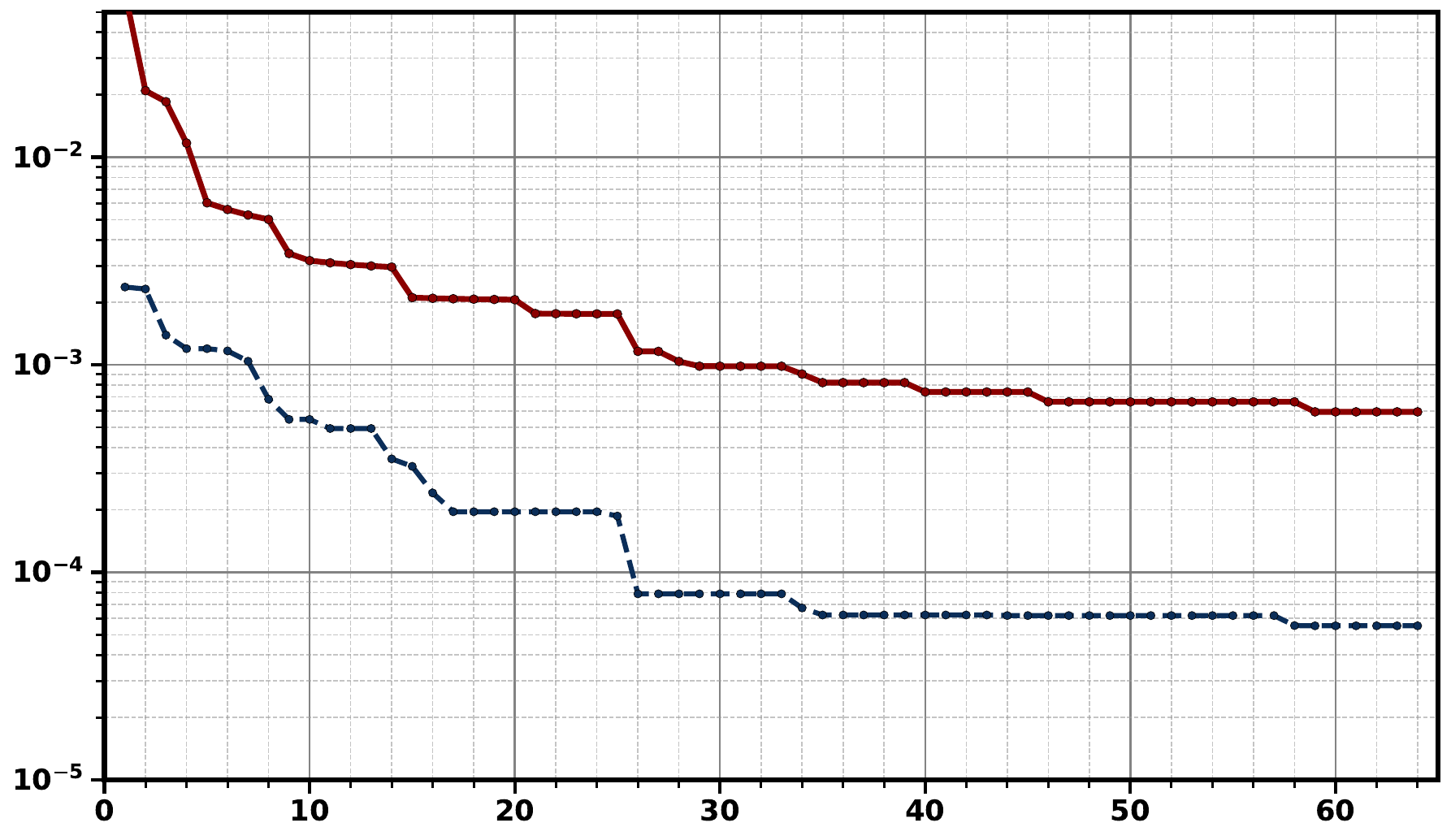}
\end{subfigure} &
\begin{subfigure}[t]{0.215\textwidth}
    \includegraphics[width=\linewidth]{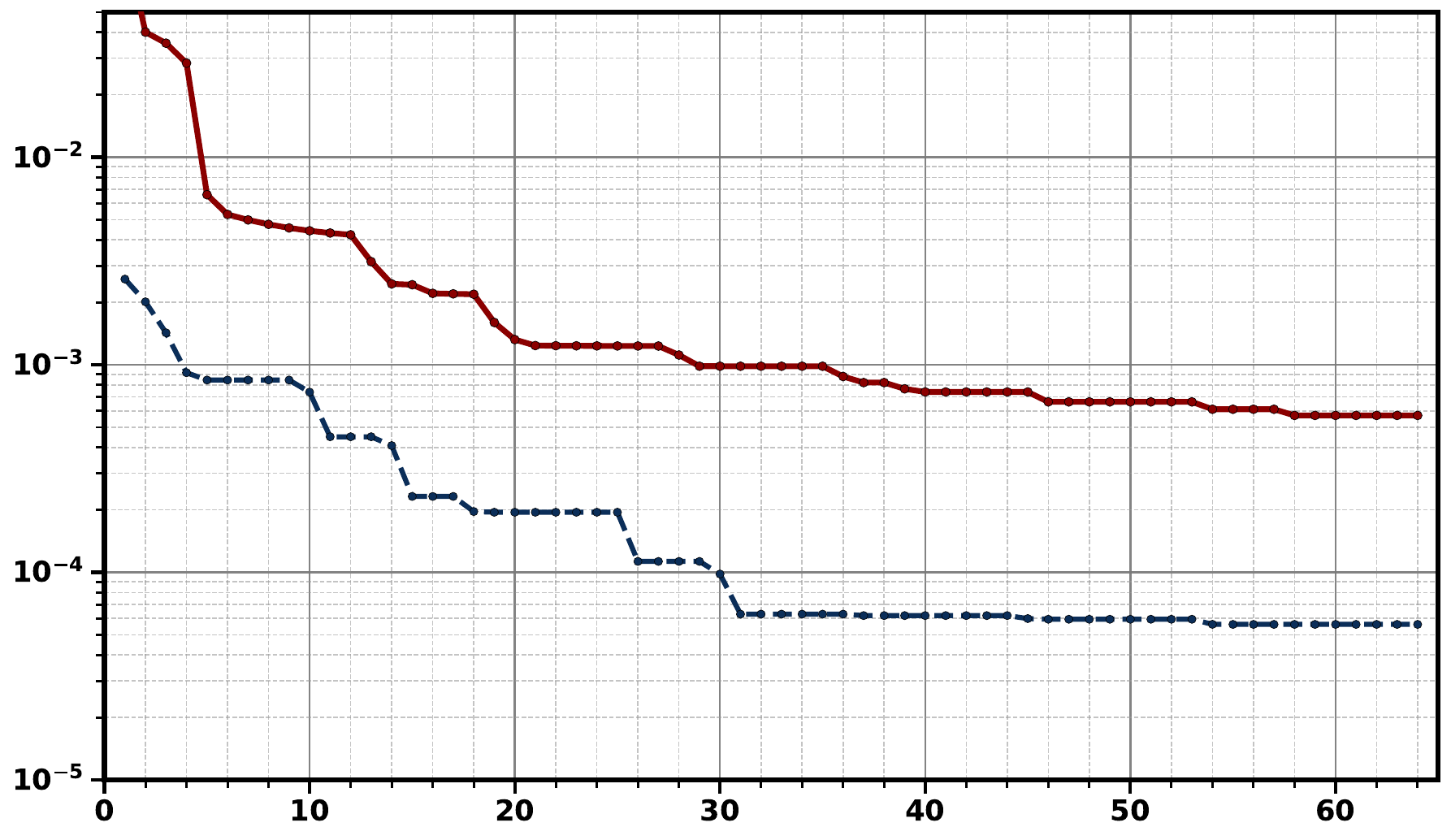}
\end{subfigure} &
\begin{subfigure}[t]{0.215\textwidth}
    \includegraphics[width=\linewidth]{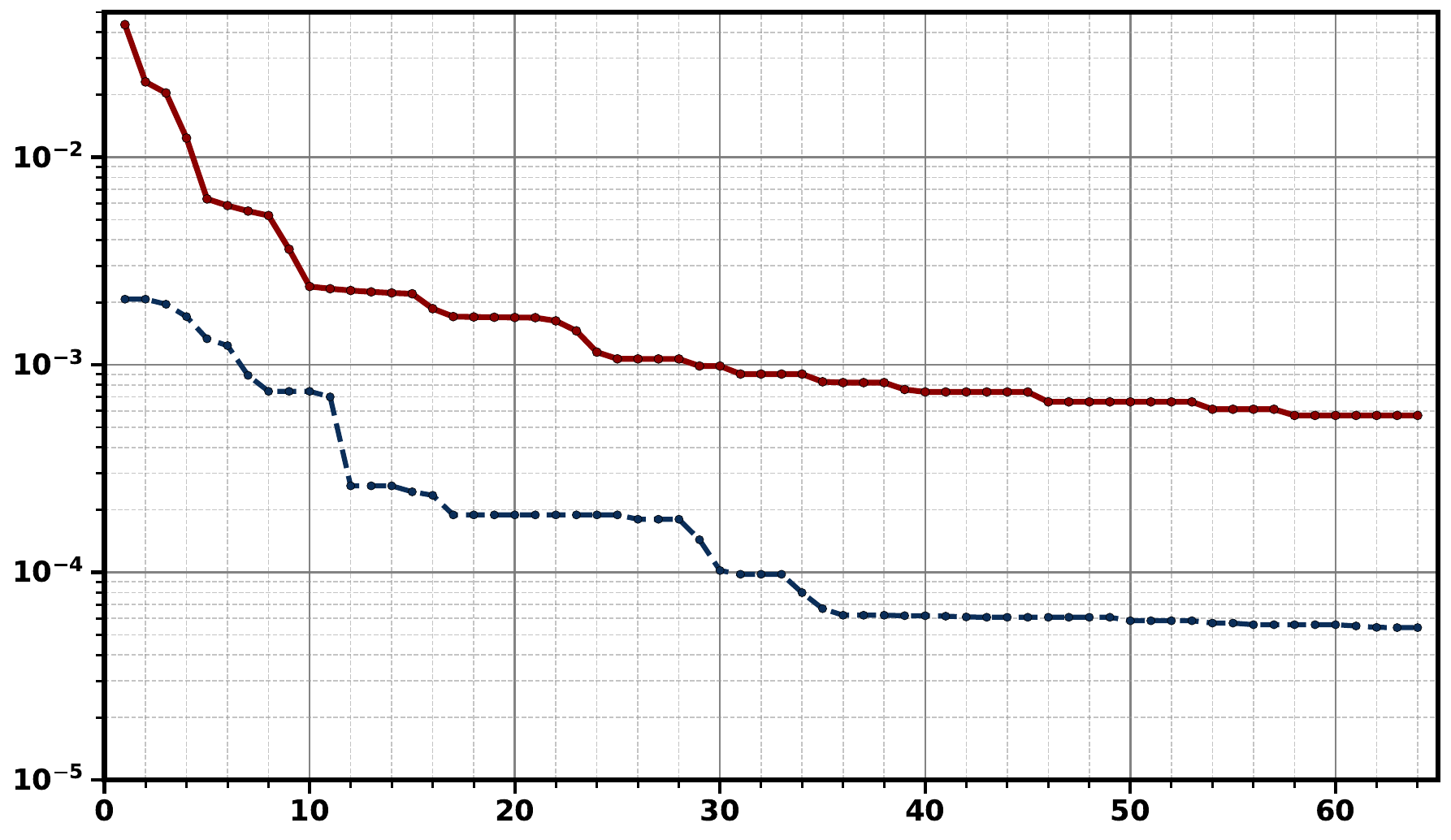}
\end{subfigure}
\end{tabular}

\vspace{0.2em}
{\footnotesize Gray dashed: early checkpoint \hspace{1.2em} Red solid: final checkpoint}

\caption{
Spectral generality across model families, optimization algorithms, and supervision regimes.
Across settings, preference tuning sharpens a leading spectral head while retaining a residual tail.
}
\label{fig:spectral_generality_grid}
\end{figure*}

\subsection{Generality Across Models and Training Regimes}
\label{sec:spectral_generality}

A single controlled trajectory could still reflect the idiosyncrasy of one model, optimizer, or supervision source. We therefore compare early and final singular spectra across 12 settings spanning model families, scales, optimization algorithms, and supervision regimes. The grid in Figure~\ref{fig:spectral_generality_grid} includes Qwen and Llama models, DPO and GRPO training, and both synthetic and benchmark-derived preference data.

Across all settings, training sharpens a small leading head while retaining a visible residual tail. The steepness of the final spectrum varies with model and regime, but the qualitative transition is stable: the final update is more organized than the early update, yet it does not collapse into only a few singular directions.

This recurrence across settings makes the spectral split a stable property of preference-induced updates rather than a visual artifact of the reference run. Preference tuning consistently reorganizes the update into a dominant head and a retained tail.

\section{Functional Separation}
\label{sec:functional_separation}

A spectral split does not by itself imply behavioral specialization. The leading directions may simply contain more update energy, while the residual directions may be too weak to matter. To test whether the head--tail structure has functional content, we convert each component back into a plug-in adapter and inject it into the same base model.

\subsection{Source-Controlled Plug-in}
\label{sec:plugin_intervention}

Programmatically verifiable tasks let us control training bias and measure solver behavior without ambiguity. We train two DPO adapters from the same base model under matched optimization settings but different training-distribution biases. \textbf{Run-Direct} emphasizes short-horizon extraction, local transformation, and shallow computation, whereas \textbf{Run-Deliberate} places more mass on tasks requiring explicit decomposition and multi-step reasoning.

For each final adapter, we split the effective update into a \textbf{spectral head} and a \textbf{residual tail} using the partition in Section~\ref{sec:spectral_head_tail}, with split rank \(k=16\). Each component is refactorized into LoRA form and loaded back into the same base model. Since head and tail come from the same checkpoint and exactly reconstruct the full update when summed, the comparison keeps the backbone, training run, and optimization history fixed. Only the injected spectral component changes.

This yields four variants on the same base model: \textbf{base}, \textbf{full}, \textbf{head-only}, and \textbf{tail-only}. All variants are evaluated on ID, OOD, and TRAP splits under the same final-answer metric. Full task construction, surface-form variants, decoding settings, and scoring details are given in Appendix~\ref{app:plugin_details}.

\begin{figure*}[htbp]
\centering
\vspace{-8pt}
\begin{subfigure}[t]{0.48\textwidth}
    \centering
    \includegraphics[width=\linewidth]{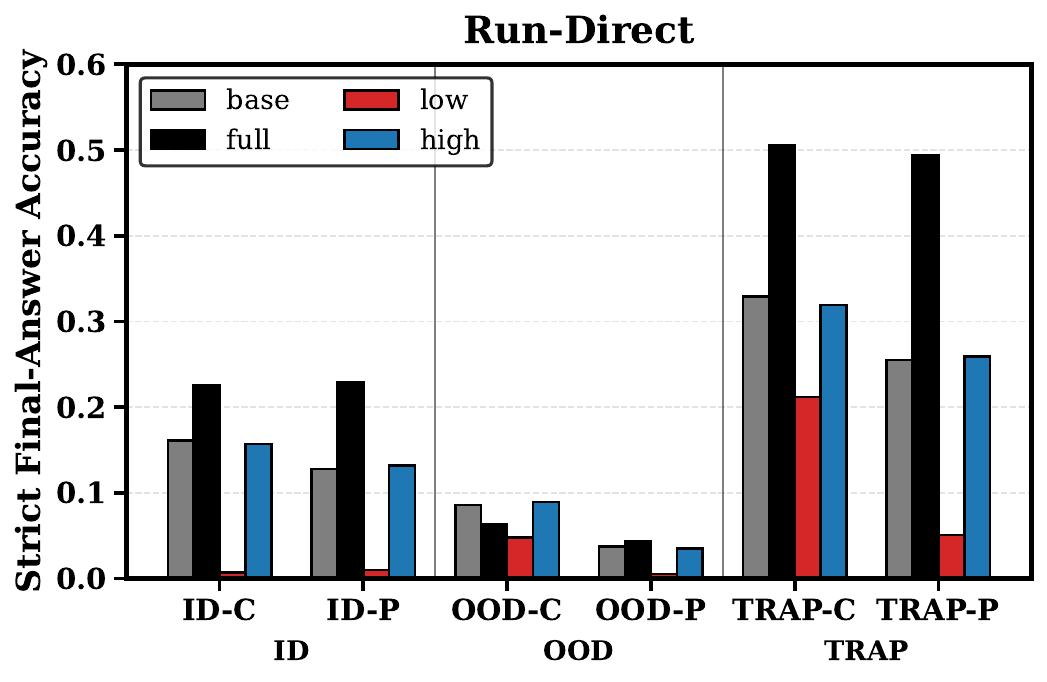}
    \caption{Run-Direct}
\end{subfigure}
\hfill
\begin{subfigure}[t]{0.48\textwidth}
    \centering
    \includegraphics[width=\linewidth]{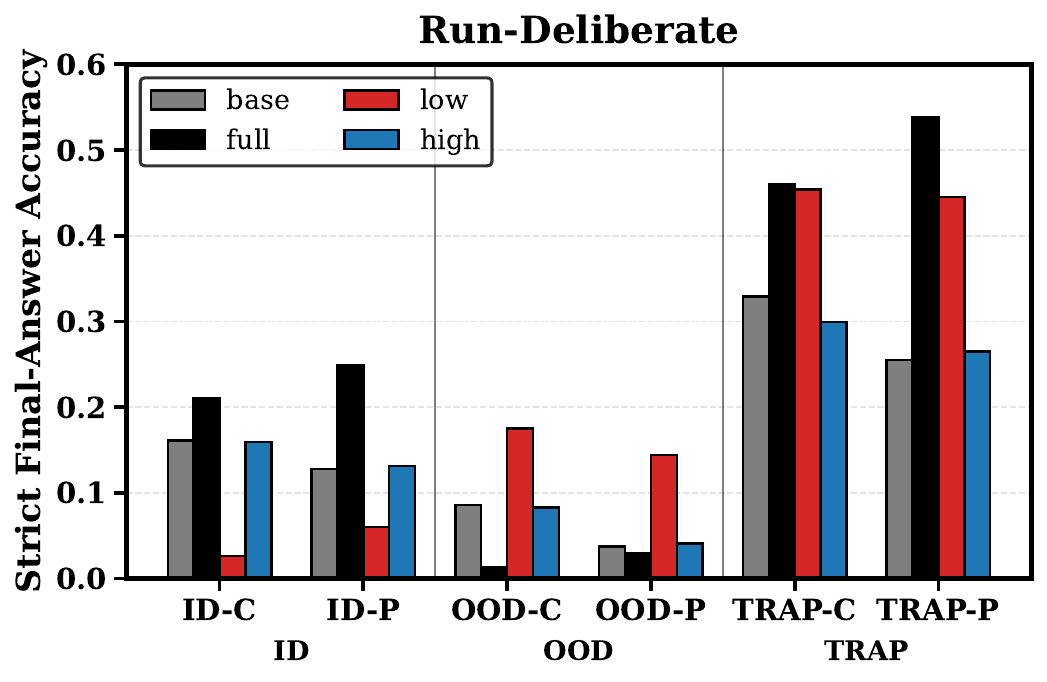}
    \caption{Run-Deliberate}
\end{subfigure}

\caption{
Source-controlled plug-in intervention at the final checkpoint with split rank \(k=16\). Each panel compares \textbf{base}, \textbf{full}, \textbf{head-only}, and \textbf{tail-only} on ID, OOD, and TRAP evaluations.
}
\label{fig:ch4_plugin_asymmetry}
\end{figure*}

\subsection{The Head Dominates Plug-in Behavior}
\label{sec:ch4_main_ablation}

Figure~\ref{fig:ch4_plugin_asymmetry} shows a clear behavioral asymmetry. Across both runs and evaluation families, \textbf{tail-only} remains close to \textbf{base}, whereas \textbf{head-only} carries the visible departure from the base model. This departure is not uniformly beneficial: in \textbf{Run-Direct}, the head is brittle and fails to recover strong OOD behavior, while in \textbf{Run-Deliberate}, it preserves a distinct profile, including stronger OOD behavior than both \textbf{base} and \textbf{full}, while remaining close to \textbf{full} on TRAP. The spectral head is therefore not merely the largest-energy fragment of the adapter; it carries the dominant endpoint effect induced by the training distribution. The scoring protocol and exact plug-in values are provided in Appendix~\ref{app:plugin_exact_values}. Plug-in isolation therefore shows that the spectral split is functional rather than merely descriptive.

\section{Deep Analysis}
\label{sec:mechanistic_analysis}

Same-run plug-in isolation shows that the spectral head and residual tail express different endpoint behaviors. This asymmetry leaves three questions unresolved. First, whether the head carries the behavioral identity of a training run, rather than only a larger share of update magnitude. Second, whether endpoint dominance makes the head sufficient during learning. Third, whether coherent preference supervision is an upstream condition for stable update formation. We address these questions through cross-run recomposition, training-time projection, and supervision corruption.

\subsection{Cross-run Composition}
\label{sec:q3_swap}

If the spectral head carries run-level solver bias, recomposing components across runs should preserve the solver profile of the head source more than that of the tail source. We test this by exchanging head and tail components between the two controlled runs.

Let \(A\) denote \textbf{Run-Direct} and \(B\) denote \textbf{Run-Deliberate}. Starting from the final checkpoints, we split each learned adapter into its spectral head and residual tail, and construct two mixed adapters:
\[
\textbf{head}(A)+\textbf{tail}(B),
\qquad
\textbf{head}(B)+\textbf{tail}(A).
\]
Recomposition is exact at the effective-update level via rank concatenation, as shown in Appendix~\ref{app:recomposition}.

\begin{figure*}[htbp]
\centering
\begin{subfigure}[t]{0.48\textwidth}
    \centering
    \includegraphics[width=\linewidth]{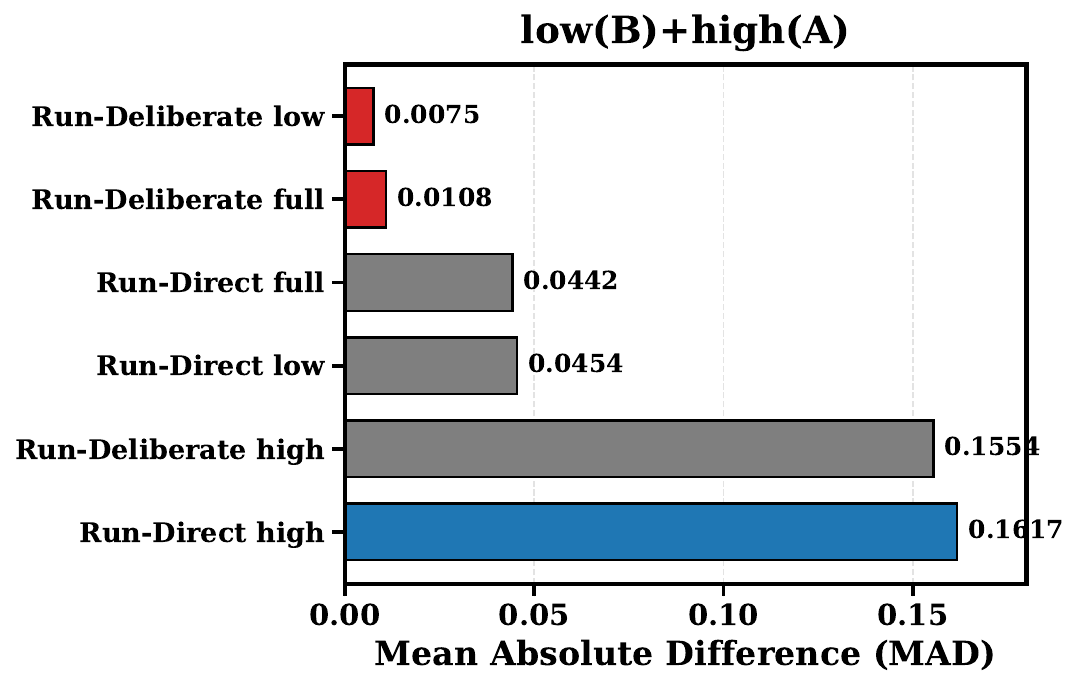}
    \caption{\(\textbf{head}(B)+\textbf{tail}(A)\)}
\end{subfigure}
\hfill
\begin{subfigure}[t]{0.48\textwidth}
    \centering
    \includegraphics[width=\linewidth]{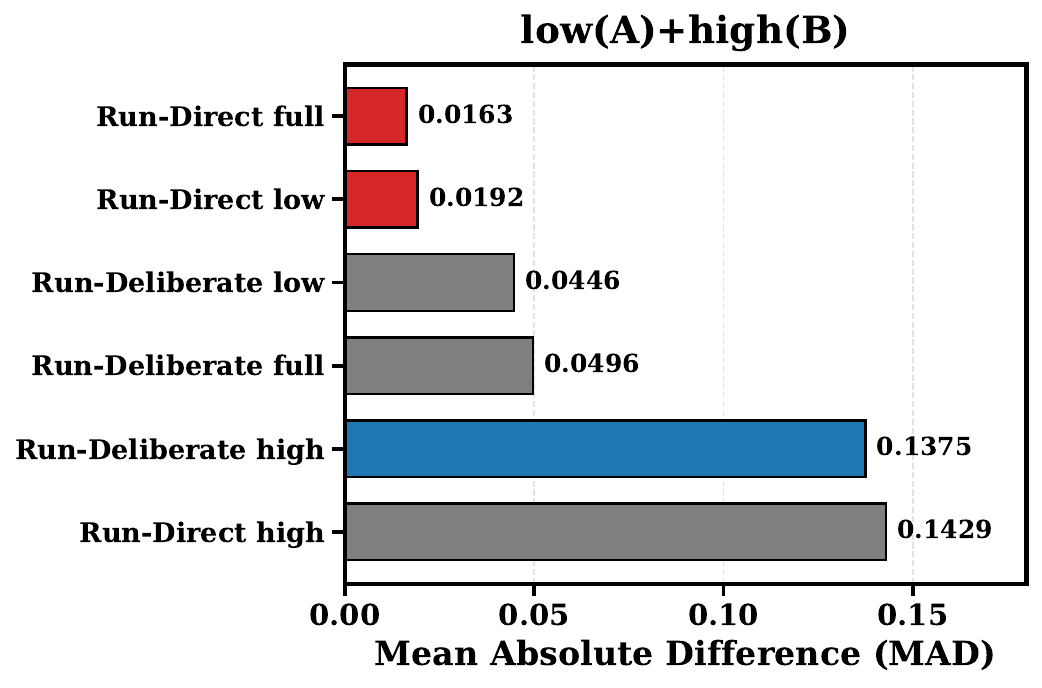}
    \caption{\(\textbf{head}(A)+\textbf{tail}(B)\)}
\end{subfigure}

\caption{Cross-run composition measured by mean absolute difference (MAD). Each panel compares one mixed adapter against candidate reference variants. Smaller values indicate greater behavioral similarity.
}
\label{fig:q3_cross_run_mad}
\end{figure*}

Figure~\ref{fig:q3_cross_run_mad} shows that the mixed adapters are closest to variants derived from their head-source runs. The adapter \(\textbf{head}(B)+\textbf{tail}(A)\) is closest to \textbf{Run-Deliberate}, whereas \(\textbf{head}(A)+\textbf{tail}(B)\) is closest to \textbf{Run-Direct}; matching the tail source alone does not determine the dominant behavior. This strengthens the plug-in result: the spectral head acts as the principal carrier of run-level solver bias, while the residual tail contributes less to the dominant similarity pattern. The recomposition procedure and MAD summaries are provided in Appendix~\ref{app:cross_run_details}.

\subsection{Training-time Spectral Projection}
\label{sec:q4_projection}

Endpoint salience and learning sufficiency are different claims. A component can dominate the behavior of a final plug-in adapter without being sufficient to recover the full solution during optimization. To separate these notions, we intervene during training by projecting each update step into one of three conditions: \textbf{full}, \textbf{head-only}, or \textbf{tail-only}.

After each optimizer step, every module-level LoRA update is projected into the selected spectral subspace. \textbf{Full} applies no projection. \textbf{Head-only} retains only the top-\(k\) singular directions, whereas \textbf{tail-only} removes them and keeps the residual component. All variants use the same base model, data, optimizer, and evaluation protocol. Implementation details are given in Appendix~\ref{app:projection_details}.

\begin{figure*}[htbp]
\centering
% \vspace{-10pt}
\begin{subfigure}[t]{0.48\textwidth}
    \centering
    \includegraphics[width=\linewidth]{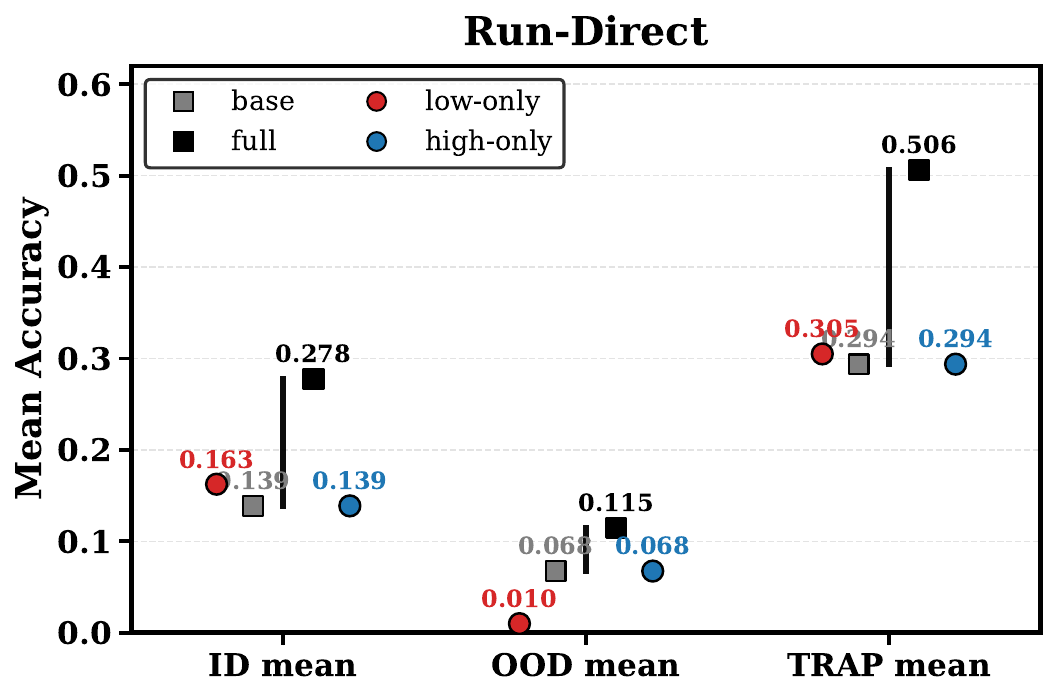}
    \caption{Run-Direct}
\end{subfigure}
\hfill
\begin{subfigure}[t]{0.48\textwidth}
    \centering
    \includegraphics[width=\linewidth]{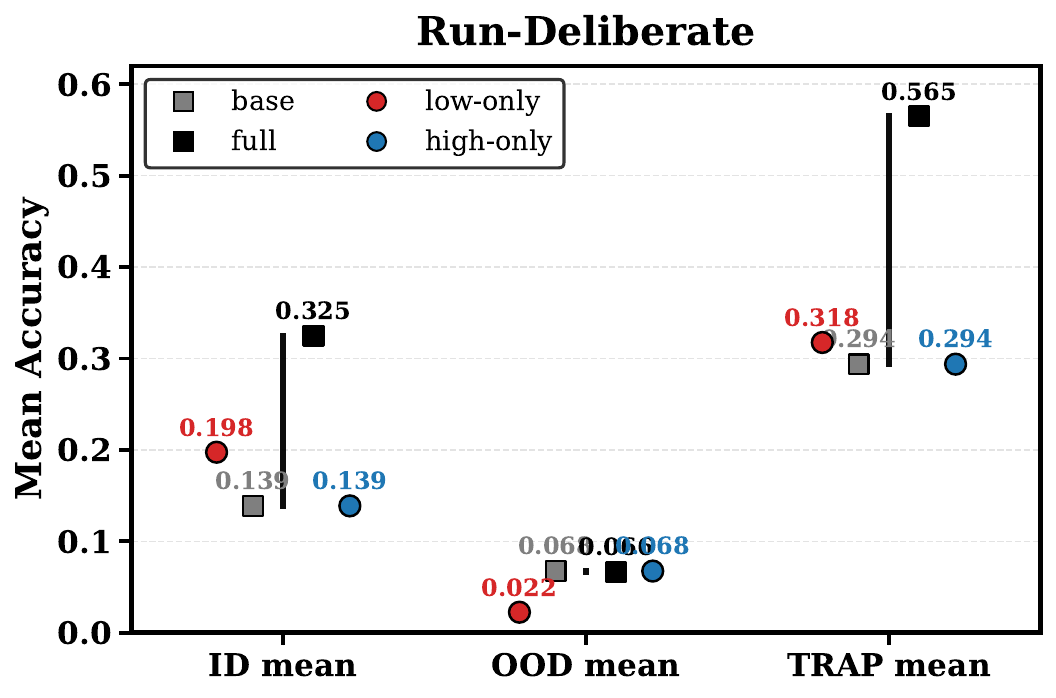}
    \caption{Run-Deliberate}
\end{subfigure}

\caption{
Training-time spectral projection. For each evaluation family, gray and black squares mark \textbf{base} and \textbf{full}, and the vertical segment indicates the full-training span. Red and blue markers denote \textbf{head-only} and \textbf{tail-only} training.
}
\label{fig:q4_projection_interval}
% \vspace{-15pt}
\end{figure*}

Figure~\ref{fig:q4_projection_interval} shows that the projected training regimes do not recover the full solution. In both runs, \textbf{tail-only} remains close to \textbf{base}, while \textbf{head-only} departs from \textbf{base} on selected ID and TRAP evaluations but remains far from \textbf{full}, especially on OOD behavior. The spectral head is therefore learning-relevant but not learning-sufficient. The residual tail is weak alone, yet the full solution is not recovered without it, suggesting a support role in preserving breadth and completing learning. The projection procedure and family-level summaries are provided in Appendix~\ref{app:projection_details}.

\subsection{Supervision Corruption}
\label{sec:q5_shuffle}

A coherent spectral structure may also depend on the coherence of the supervision that drives the update. We test this upstream condition by corrupting prompt--preference alignment while keeping the prompts fixed. Specifically, we corrupt a fraction \(\rho \in \{0.25,0.50,1.00\}\) of the training set with \texttt{shuffle\_pairs}, replacing each corrupted sample's preference pair with that of another example. Details of the corruption procedure are given in Appendix~\ref{app:shuffle_details}.

\begin{figure*}[htbp]
\centering
% \vspace{-10pt}
\begin{subfigure}[t]{0.48\textwidth}
    \centering
    \includegraphics[width=\linewidth]{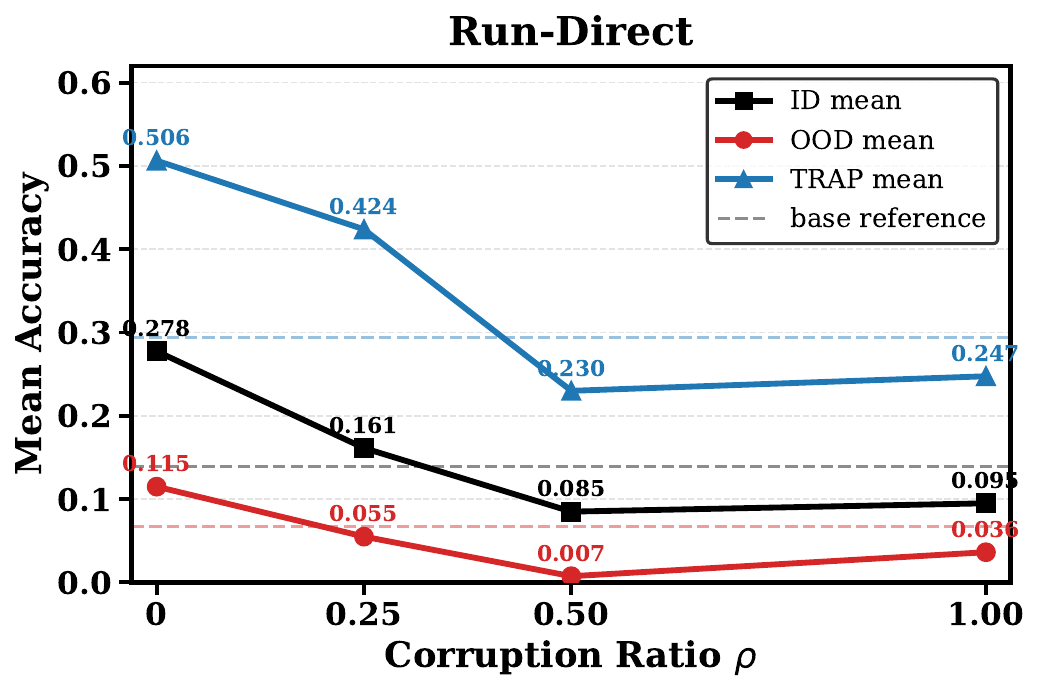}
    \caption{Run-Direct}
\end{subfigure}
\hfill
\begin{subfigure}[t]{0.48\textwidth}
    \centering
    \includegraphics[width=\linewidth]{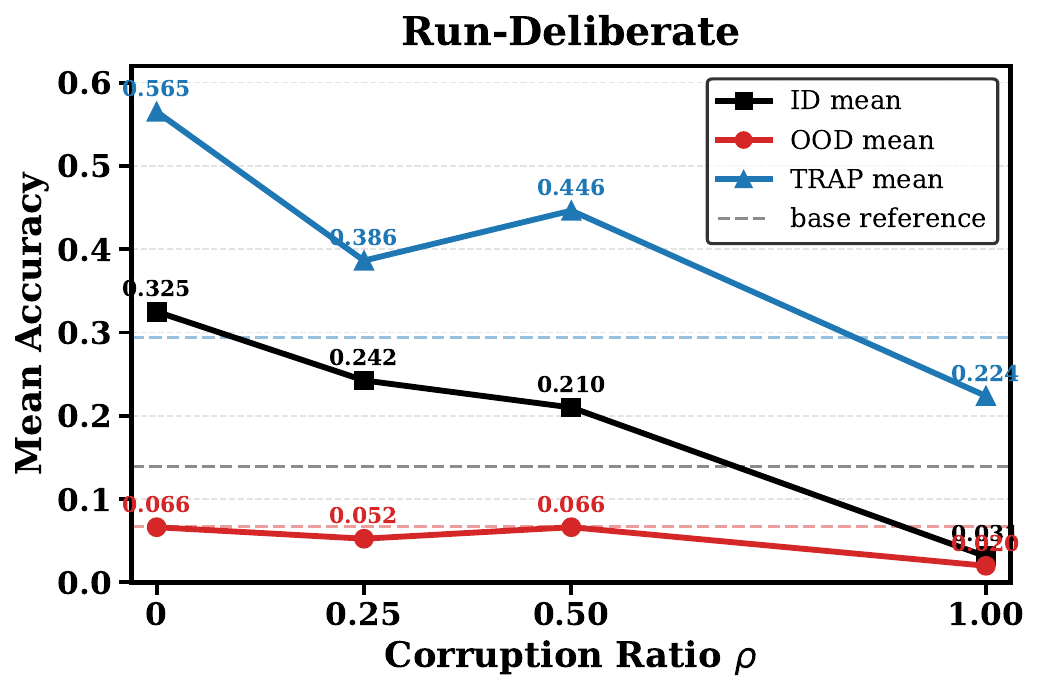}
    \caption{Run-Deliberate}
\end{subfigure}

\caption{
Supervision corruption under \texttt{shuffle\_pairs}. Each panel reports family-level mean accuracy across corruption ratios \(\rho\), where \(\rho=0\) denotes uncorrupted training. Dashed lines mark base-model performance.
}
\label{fig:q5_shuffle_curves}
% \vspace{-10pt}
\end{figure*}

Figure~\ref{fig:q5_shuffle_curves} shows that full corruption sharply degrades both runs, especially on ID and OOD behavior. Stable preference learning thus depends on preserving prompt--preference consistency. Partial corruption is not a simple linear noise response. In \textbf{Run-Direct}, moderate corruption already causes substantial degradation, with \(\rho=0.50\) approaching collapse across evaluations. In \textbf{Run-Deliberate}, intermediate corruption levels retain substantial ID and TRAP gains over \textbf{base}, even as OOD transfer weakens.

These results suggest that prompt--preference consistency is not merely a label-quality detail. It is a plausible upstream condition for coherent functional separation. The evidence, however, is behavioral rather than spectrally closed. It supports the supervision-consistency hypothesis, but does not establish how corruption changes head strength, tail thickness, or spectral trajectories. The corruption procedure and family-level summaries are provided in Appendix~\ref{app:shuffle_details}.

\section{Discussion and Limitations}
\label{sec:discussion}

\subsection{From Spectral Organization to Post-Training Trade-offs}
\label{sec:disc_tradeoffs}

\textbf{Signal coherence shapes update geometry.}
The head--tail organization suggests that preference tuning does not distribute update mass uniformly. A plausible explanation is that globally coherent directions, shared across prompts, minibatches, and preference comparisons, accumulate into the spectral head. More conditional, weakly reinforced, or locally conflicting directions align less globally and remain dispersed in the residual tail. The supervision-corruption results are consistent with this view: breaking prompt--preference alignment injects incompatible pressures and weakens stable preference learning. We do not claim a closed spectral mechanism, but the evidence supports a link between coherent supervision and coherent update geometry.

\noindent\textbf{Endpoint salience is not learning sufficiency.}
A component that controls final plug-in behavior need not be sufficient for learning. The spectral head dominates endpoint behavior and cross-run recomposition, yet head-only training fails to recover the full solution, especially on OOD behavior. The residual tail is weak alone, but the full solution is not recovered without it. This separates behavioral payload from optimization support: some directions matter less by visibly determining the endpoint, and more by helping preserve breadth and complete learning.

\noindent\textbf{Alignment gain and coverage loss may share a structural source.}
Preference tuning consolidates coherent, reward-consistent directions into the spectral head. The same process can underrepresent behavior that is conditional, long-tail, or inconsistently reinforced. From this perspective, alignment gain and coverage loss are not only consequences of reward design or data distribution. They can also arise from how the learned update organizes behavioral change.

% \noindent\textbf{Broader Impact.}
% These findings suggest that post-training should monitor update geometry alongside reward and task metrics. Head concentration, tail retention, and effective-rank dynamics may help diagnose over-consolidation of dominant behavior at the expense of conditional breadth. Such diagnostics can improve transparency and reliability, but spectral interventions could also be misused to selectively alter safety-relevant behavior. We therefore view spectrum-aware analysis as a tool for auditing and improving post-training, not for bypassing safety constraints.

\subsection{Limitations}
\label{sec:limitations}

\noindent\textbf{Update-level analysis.}
Our analysis identifies structure in learned updates, not the circuits that implement the resulting behavior. It shows that spectral components have different behavioral roles, but does not locate the specific attention heads, MLP features, or computational pathways involved.

\noindent\textbf{Incomplete supervision-to-spectrum link.}
The corruption experiments show that prompt--preference consistency affects stable preference learning, but they do not directly track how corruption changes head strength, tail thickness, subspace stability, or effective-rank trajectories.

\noindent\textbf{Controlled intervention setting.}
Our interventions rely on LoRA-style updates with a frozen backbone, where effective updates are exactly decomposable and can be refactorized into plug-in adapters. This enables controlled causal tests, but leaves open how fully the same structure transfers to full-parameter finetuning, larger RLHF pipelines, proprietary systems, and more open-ended instruction-following settings.

\section{Conclusion}
\label{sec:con}

We studied preference post-training as structured update reorganization. Preference-induced updates consistently form a spectral head--tail organization, where a compact head carries dominant endpoint behavior and run-level solver bias, while a retained residual tail is weak alone but appears necessary for learning completeness and behavioral breadth.
This perspective reframes alignment gain and coverage loss as consequences of how the update itself is organized. Preference tuning consolidates coherent behavioral directions, but can leave conditional or weakly reinforced behavior fragile. Future post-training should therefore monitor update geometry alongside endpoint rewards. Spectrum-aware diagnostics may help improve alignment without unnecessarily narrowing behavioral coverage.

%--------------------------------------------------------------
%     Bibliography
%--------------------------------------------------------------
\newpage
\bibliography{ref}
\bibliographystyle{abbrvnat}

\newpage
\appendix
\onecolumn
\section{Update Construction and Spectral Operations}
\label{app:update_details}

All spectral operations in the paper are defined on the scaled effective update, not on the LoRA factors separately. This convention makes the decomposition independent of a particular adapter parameterization and makes component isolation, recomposition, and projection exact at the effective-update level. This appendix gives the implementation details for constructing these updates, splitting them spectrally, converting components back into adapter form, and computing spectral statistics.

\subsection{Targeted Modules and Update Units}
\label{app:target_modules}

We analyze preference-induced updates at the level of module-wise linear projections inside Transformer blocks. For the decoder-only architectures studied in the paper, the targeted module set is
\[
\mathcal M =
\{
\texttt{q\_proj},
\texttt{k\_proj},
\texttt{v\_proj},
\texttt{o\_proj},
\texttt{gate\_proj},
\texttt{up\_proj},
\texttt{down\_proj}
\}.
\]
For a model with \(L\) Transformer blocks, the checkpoint-level update is represented as
\[
\Delta \theta^{(t)}
=
\{
\Delta W_{\ell,j}^{(t)}
\}_{\ell=1,\dots,L,\; j\in\mathcal M}.
\]
We also use the block-restricted collection
\[
\Delta \theta_{\ell}^{(t)}
=
\{
\Delta W_{\ell,j}^{(t)}
\}_{j\in\mathcal M}
\]
when summarizing updates within a Transformer block. Unless otherwise stated, all operations below are applied independently to each module-level update \(\Delta W_{\ell,j}^{(t)}\).

\subsection{Effective Updates under Dense Tuning and LoRA}
\label{app:effective_update_lora}

Let \(W_{\ell,j}^{(0)}\) denote the pre-tuning weight of module \(j\) in block \(\ell\), and let \(W_{\ell,j}^{(t)}\) denote the corresponding weight at checkpoint \(t\). The effective update is the additive patch satisfying
\[
W_{\ell,j}^{(t)}
=
W_{\ell,j}^{(0)}
+
\Delta W_{\ell,j}^{(t)}.
\]
For dense finetuning, this gives
\[
\Delta W_{\ell,j}^{(t)}
=
W_{\ell,j}^{(t)}
-
W_{\ell,j}^{(0)}.
\]

In the LoRA setting, the backbone is frozen and the learned change is represented by low-rank adapter factors. For a target module with input dimension \(d_{\mathrm{in}}\), output dimension \(d_{\mathrm{out}}\), and LoRA rank \(r_0\), we write
\[
A_{\ell,j}^{(t)}
\in
\mathbb R^{r_0\times d_{\mathrm{in}}},
\qquad
B_{\ell,j}^{(t)}
\in
\mathbb R^{d_{\mathrm{out}}\times r_0}.
\]
The scaled effective update is
\[
\Delta W_{\ell,j}^{(t)}
=
s_{\ell,j}^{(t)}
B_{\ell,j}^{(t)}
A_{\ell,j}^{(t)},
\]
where \(s_{\ell,j}^{(t)}\) absorbs implementation-specific scaling conventions, including the usual LoRA factor \(\alpha/r_0\). All spectral decompositions, statistics, and interventions are applied to this scaled matrix. We do not decompose \(A_{\ell,j}^{(t)}\) or \(B_{\ell,j}^{(t)}\) separately.

Since \(\operatorname{rank}(\Delta W_{\ell,j}^{(t)})\le r_0\), each LoRA update has at most \(r_0\) nonzero singular values. Unless otherwise stated, the main experiments use \(r_0=64\).

\subsection{Spectral Decomposition}
\label{app:svd_computation}

For each module-level effective update, we compute
\[
\Delta W_{\ell,j}^{(t)}
=
U_{\ell,j}^{(t)}
\Sigma_{\ell,j}^{(t)}
\left(
V_{\ell,j}^{(t)}
\right)^{\top},
\]
with singular values sorted in non-increasing order. The decomposition can be obtained either by materializing \(\Delta W_{\ell,j}^{(t)}\) and applying an economy SVD, or by using an equivalent compact low-rank computation. Both procedures define the same spectral components at the level of the scaled effective update.

All reported spectra and effective-rank statistics are computed after applying the LoRA scaling factor \(s_{\ell,j}^{(t)}\).

\subsection{Spectral Head and Residual Tail}
\label{app:head_tail_split}

Given a split rank \(r\), the spectral head is the leading truncated component
\[
\Delta W_{\ell,j}^{\mathrm{head}}(r)
=
U_{\ell,j}^{(:,1:r)}
\Sigma_{\ell,j}^{(1:r,1:r)}
\left(
V_{\ell,j}^{(:,1:r)}
\right)^{\top}.
\]
The residual tail is the complementary component
\[
\Delta W_{\ell,j}^{\mathrm{tail}}(r)
=
\Delta W_{\ell,j}
-
\Delta W_{\ell,j}^{\mathrm{head}}(r).
\]
Thus,
\[
\Delta W_{\ell,j}
=
\Delta W_{\ell,j}^{\mathrm{head}}(r)
+
\Delta W_{\ell,j}^{\mathrm{tail}}(r).
\]
Unless otherwise stated, the plug-in, recomposition, and projection experiments use split rank \(r=16\) with LoRA rank \(r_0=64\).

\subsection{Refactorizing Spectral Components into LoRA Form}
\label{app:svd_to_lora}

Intervention requires each spectral component to be loaded as an adapter. Given any rank-\(q\) update component
\[
\Delta W_c
=
U_c \Sigma_c V_c^\top,
\]
we construct LoRA-style factors
\[
B_c
=
U_c \Sigma_c^{1/2},
\qquad
A_c
=
\Sigma_c^{1/2} V_c^\top,
\]
so that
\[
B_c A_c
=
\Delta W_c.
\]
The refactorized adapter uses scaling factor \(1\), since the original LoRA scaling has already been absorbed into \(\Delta W_c\).

If an implementation requires a fixed adapter rank, components with rank below the required size are padded with zero columns in \(B_c\) and zero rows in \(A_c\). This padding preserves the effective update exactly.

\subsection{Isolation and Recomposition}
\label{app:recomposition}

The additive update view makes component isolation and recomposition exact at the effective-update level.

\paragraph{Isolation.}
For a targeted module, isolating a component means loading only that component on top of the frozen base weight:
\[
W_{\ell,j}
=
W_{\ell,j}^{(0)}
+
\Delta W_{\ell,j}^{\mathrm{head}}(r)
\]
or
\[
W_{\ell,j}
=
W_{\ell,j}^{(0)}
+
\Delta W_{\ell,j}^{\mathrm{tail}}(r).
\]
The same operation is applied independently to every targeted module.

\paragraph{Cross-run recomposition.}
Let \(A\) and \(B\) denote two independently trained runs. A mixed update such as
\[
\Delta W_{\mathrm{mix}}
=
\Delta W_A^{\mathrm{head}}(r)
+
\Delta W_B^{\mathrm{tail}}(r)
\]
can be represented exactly by rank concatenation. If
\[
\Delta W_A^{\mathrm{head}}
=
B_A^{h} A_A^{h},
\qquad
\Delta W_B^{\mathrm{tail}}
=
B_B^{t} A_B^{t},
\]
then
\[
\Delta W_{\mathrm{mix}}
=
\begin{bmatrix}
B_A^{h} & B_B^{t}
\end{bmatrix}
\begin{bmatrix}
A_A^{h} \\
A_B^{t}
\end{bmatrix}.
\]
This construction preserves the exact sum of the two effective-update components.

\subsection{Training-Time Spectral Projection}
\label{app:projection}

For training-time projection, the effective update is constrained after each optimizer step. Let \(\Delta W^{(\tau)}\) denote the current module-level update at optimizer step \(\tau\). We replace it with one of
\[
\Pi_{\mathrm{full}}(\Delta W^{(\tau)})
=
\Delta W^{(\tau)},
\]
\[
\Pi_{\mathrm{head}}(\Delta W^{(\tau)})
=
\Delta W^{\mathrm{head}}(r),
\]
\[
\Pi_{\mathrm{tail}}(\Delta W^{(\tau)})
=
\Delta W^{\mathrm{tail}}(r).
\]
The projected update is refactorized into LoRA form using Appendix~\ref{app:svd_to_lora} and written back to the adapter parameters before the next forward pass. Projection is applied independently to each targeted module.

\subsection{Entropy Effective Rank}
\label{app:effective_rank_details}

For singular values \(\{\sigma_i\}\), define
\[
p_i
=
\frac{\sigma_i}{\sum_k \sigma_k}.
\]
The spectral entropy is
\[
H
=
-\sum_i p_i\log(p_i+\varepsilon),
\]
and the effective rank is
\[
r_{\mathrm{eff}}
=
\exp(H).
\]
Smaller \(r_{\mathrm{eff}}\) indicates concentration in fewer dominant singular directions, while larger \(r_{\mathrm{eff}}\) indicates a flatter spectrum. Unless otherwise stated, checkpoint-level summaries report the median effective rank over all targeted modules:
\[
\widetilde r_{\mathrm{eff}}^{(t)}
=
\operatorname{median}_{\ell,j}
\;
r_{\mathrm{eff},\ell,j}^{(t)}.
\]

\subsection{Numerical Safeguards}
\label{app:numerical_safeguards}

All spectra are computed after forming the scaled effective update. Singular values below a fixed numerical tolerance are treated as negligible when computing quantities that depend on rank support. For entropy effective rank, the normalization \(p_i=\sigma_i/\sum_k\sigma_k\) is applied only when the denominator is nonzero, and a small constant \(\varepsilon\) is used inside the logarithm. The same tolerance and entropy convention are used across all compared runs.

\section{Details for Spectral Formation}
\label{app:spectral_full}

This appendix records the experimental grid and numerical summaries used for the spectral-formation analysis in Section~\ref{sec:spectral_separation}. All spectra are computed from the scaled effective updates defined in Appendix~\ref{app:effective_update_lora}, and checkpoint-level summaries aggregate module-level quantities over the targeted projections in Appendix~\ref{app:target_modules}.

\subsection{Experimental Setup}
\label{app:spectral_setup}

\begin{table}[h]
\centering
\caption{Experimental design for tracking spectral formation over training.}
\label{tab:spectral_setup_full}
\begin{tabular}{ll}
\hline
Component & Setting \\
\hline
Reference trajectory & Qwen2.5-0.5B with controlled synthetic supervision \\
Generality models & Qwen3-1.7B, Qwen3-8B, Llama-3.1-8B \\
Algorithms & DPO, GRPO \\
Supervision regimes & Controlled synthetic preferences; benchmark-derived preference data \\
Training form & LoRA-only updates with frozen backbone \\
LoRA rank & \(r_0=64\), unless otherwise specified \\
Target modules & Attention and MLP projections listed in Appendix~\ref{app:target_modules} \\
Main observables & Checkpointed singular spectra; effective-rank trajectories \\
\hline
\end{tabular}
\end{table}

For each saved checkpoint \(t\), we construct the module-level effective update \(\Delta W_{\ell,j}^{(t)}\), compute its singular-value spectrum, and summarize spectral concentration with entropy effective rank. Unless otherwise stated, run-level effective-rank summaries are reported as the median over all targeted modules.

\subsection{Reference-Trajectory Effective Rank}
\label{app:reference_effective_rank}

In the reference trajectory, the median effective rank follows a non-monotonic pattern. It decreases from \(43.57\) at step 31 to \(41.48\) at step 93, indicating early concentration of update mass into fewer leading directions. It then rebounds to \(42.60\) by step 217 and ends at \(42.03\) by step 837.

The rebound after the early contraction is the relevant feature. A monotonic low-rank-collapse account would predict continued decrease as training proceeds. Instead, the trajectory shows early concentration followed by stabilization, consistent with the coexistence of a growing spectral head and a retained residual tail.

\subsection{Reference-Trajectory Singular Spectra}
\label{app:reference_spectra_details}

The raw singular spectra show the same formation process directly. At step 31, the largest singular values are approximately \(0.0014\), \(0.0012\), and \(0.0007\), while the remaining singular values lie near the \(10^{-4}\) scale. By step 837, the leading singular values increase to approximately \(0.0170\), \(0.0069\), and \(0.0059\).

Thus, training substantially amplifies the leading directions, producing a clearer spectral head. At the same time, the residual spectrum remains visible rather than disappearing. This combination explains why effective rank contracts early but does not continue collapsing throughout training.

\subsection{Generality Grid}
\label{app:spectral_generality_details}

The generality grid uses the same early-versus-final comparison protocol across 12 model--algorithm--supervision settings. Each panel compares the singular spectrum of an early checkpoint with that of the final checkpoint for one setting.

Across the grid, the final spectrum consistently exhibits a more pronounced leading head than the early spectrum, while retaining a visible residual tail. The slope and separation strength vary across models and regimes, but the qualitative transition remains the same. This supports the interpretation that head--tail organization is not tied to the reference trajectory alone.

\section{Details for Functional Separation}
\label{app:plugin_details}

The plug-in experiments isolate the behavioral effect of spectral components under a fixed source. The base model, training run, checkpoint, decoding protocol, and scoring rule are held fixed. Only the loaded update component changes. This appendix specifies the controlled task setting, the two training-distribution biases, the plug-in construction, and the exact values used in Figure~\ref{fig:ch4_plugin_asymmetry}.

\subsection{Controlled Task Setting}
\label{app:plugin_task_setting}

We use synthetic, programmatically verifiable tasks as a controlled intervention testbed. Their role is to provide an environment where training pressure, solver bias, and evaluation targets can be specified precisely. This makes component-level intervention easier to interpret than in open-ended generation tasks.

All tasks follow the same output protocol. The model is prompted to solve the task and produce a final answer after a fixed marker \(\texttt{FINAL:}\). Only the content following this marker is used for scoring.

\subsection{Training Runs}
\label{app:plugin_training_runs}

The plug-in comparison uses two controlled DPO runs. Both runs start from the same base model and share the same formatting protocol, optimizer configuration, decoding setup, and evaluation procedure. They differ in training-distribution bias, which creates two distinct solver pressures under otherwise matched conditions.

\paragraph{Run-Direct.}
\textbf{Run-Direct} is biased toward short-horizon tasks solved by direct extraction, local transformation, or shallow computation.

\paragraph{Run-Deliberate.}
\textbf{Run-Deliberate} assigns more probability mass to tasks requiring explicit intermediate decomposition and multi-step reasoning.

\subsection{Plug-in Construction}
\label{app:plugin_construction}

For each final adapter, we split every module-level effective update into a spectral head and residual tail using Appendix~\ref{app:head_tail_split}. Unless otherwise stated, the split rank is \(k=16\).

For each run, we evaluate four variants:
\begin{itemize}[leftmargin=*,noitemsep,topsep=0pt]
    \item \textbf{base}: the original base model with no adapter loaded;
    \item \textbf{full}: the full learned LoRA adapter;
    \item \textbf{head-only}: only the spectral head is loaded;
    \item \textbf{tail-only}: only the residual tail is loaded.
\end{itemize}

The \textbf{head-only} and \textbf{tail-only} adapters are obtained by the refactorization procedure in Appendix~\ref{app:svd_to_lora}. Since the head and tail come from the same checkpoint and exactly reconstruct the full effective update when summed, the comparison controls for the backbone, training run, checkpoint, and optimization history. Only the injected spectral component differs.

\subsection{Evaluation and Scoring}
\label{app:plugin_eval_scoring}

We evaluate each variant on ID, OOD, and TRAP splits. ID follows the same distributional bias as the corresponding training run. OOD contains tasks requiring stronger transfer beyond the dominant training bias. TRAP is designed to expose shortcutting, brittle pattern matching, or over-deliberate solving.

Each split is evaluated under both Canonical and Perturbed surface forms. Canonical follows the standard task wording, while Perturbed changes superficial phrasing without changing the answer.

All comparisons use strict final-answer exact match. Only the text after \(\texttt{FINAL:}\) is extracted and compared with the programmatically computed ground truth. Decoding is greedy and identical across variants.

\subsection{Exact Plug-in Results}
\label{app:plugin_exact_values}

Table~\ref{tab:ablation_ckpt950} reports the exact plug-in values corresponding to Figure~\ref{fig:ch4_plugin_asymmetry}. These values support the main-text observation that \textbf{tail-only} remains close to \textbf{base}, while \textbf{head-only} carries the main visible departure from the base model, with a run-dependent behavioral profile.

\begin{table}[t]
\centering
\caption{Source-controlled plug-in results at the final checkpoint with split rank \(k=16\). Metric is strict final-answer accuracy under greedy decoding. ID-C/P, OOD-C/P, and TRAP-C/P denote the Canonical and Perturbed surface forms of the corresponding evaluation splits.}
\label{tab:ablation_ckpt950}
\begin{tabular}{llcccccc}
\hline
Run & Variant & ID-C & ID-P & OOD-C & OOD-P & TRAP-C & TRAP-P \\
\hline
Run-Direct & base & 0.161 & 0.128 & 0.086 & 0.037 & 0.329 & 0.255 \\
Run-Direct & full & 0.226 & 0.229 & 0.063 & 0.043 & 0.506 & 0.494 \\
Run-Direct & head-only  & 0.007 & 0.010 & 0.048 & 0.005 & 0.212 & 0.051 \\
Run-Direct & tail-only & 0.157 & 0.132 & 0.089 & 0.035 & 0.319 & 0.259 \\
\hline
Run-Deliberate & base & 0.161 & 0.128 & 0.086 & 0.037 & 0.329 & 0.255 \\
Run-Deliberate & full & 0.210 & 0.249 & 0.013 & 0.029 & 0.460 & 0.538 \\
Run-Deliberate & head-only  & 0.026 & 0.060 & 0.175 & 0.144 & 0.454 & 0.445 \\
Run-Deliberate & tail-only & 0.159 & 0.131 & 0.083 & 0.041 & 0.299 & 0.265 \\
\hline
\end{tabular}
\end{table}

\section{Details for Deep Analysis}
\label{app:mechanistic_details}

This appendix provides the details omitted from Section~\ref{sec:mechanistic_analysis}. It covers cross-run composition, training-time spectral projection, and supervision corruption.

\subsection{Cross-run Composition}
\label{app:cross_run_details}

Let \(A\) denote \textbf{Run-Direct} and \(B\) denote \textbf{Run-Deliberate}. Starting from the final checkpoints, we split each learned adapter into a spectral head and residual tail, then construct
\[
\mathrm{head}(A)+\mathrm{tail}(B),
\qquad
\mathrm{head}(B)+\mathrm{tail}(A).
\]
The recomposition is exact at the effective-update level using the rank-concatenation procedure in Appendix~\ref{app:recomposition}.

To quantify behavioral similarity, we use the mean absolute difference (MAD) across the six evaluation metrics:
\[
\mathrm{MAD}(u,v)
=
\frac{1}{6}
\sum_{m \in \{\mathrm{ID\text{-}C,ID\text{-}P,OOD\text{-}C,OOD\text{-}P,TRAP\text{-}C,TRAP\text{-}P}\}}
|u_m - v_m|.
\]
Smaller MAD indicates greater behavioral similarity.

\begin{table}[t]
\centering
\caption{MAD-based similarity summary for within-run and cross-run comparisons. Smaller values indicate greater behavioral similarity.}
\label{tab:q3_cross_run_delta}
\begin{tabular}{llc}
\hline
Comparison & Compared to & MAD \\
\hline
Within-run & Run-Direct: full vs head-only & 0.2047 \\
Within-run & Run-Direct: tail-only vs base & 0.0045 \\
Within-run & Run-Deliberate: full vs head-only & 0.1248 \\
Within-run & Run-Deliberate: tail-only vs base & 0.0087 \\
\hline
Cross-run & head(B)+tail(A) vs Run-Deliberate head-only & 0.0320 \\
Cross-run & head(B)+tail(A) vs Run-Deliberate full & 0.1080 \\
Cross-run & head(B)+tail(A) vs Run-Direct head-only & 0.1480 \\
Cross-run & head(B)+tail(A) vs Run-Direct full & 0.1330 \\
Cross-run & head(B)+tail(A) vs Run-Deliberate tail-only & 0.1190 \\
Cross-run & head(B)+tail(A) vs Run-Direct tail-only & 0.1260 \\
\hline
Cross-run & head(A)+tail(B) vs Run-Direct head-only & 0.0340 \\
Cross-run & head(A)+tail(B) vs Run-Direct full & 0.1830 \\
Cross-run & head(A)+tail(B) vs Run-Deliberate head-only & 0.1450 \\
Cross-run & head(A)+tail(B) vs Run-Deliberate full & 0.1870 \\
Cross-run & head(A)+tail(B) vs Run-Direct tail-only & 0.1040 \\
Cross-run & head(A)+tail(B) vs Run-Deliberate tail-only & 0.1120 \\
\hline
\end{tabular}
\end{table}

\subsection{Training-time Spectral Projection}
\label{app:projection_details}

To test whether the spectral head is sufficient during learning, we intervene after each optimizer step by projecting every targeted module into one of three variants: \textbf{full}, \textbf{head-only}, or \textbf{tail-only}. The \textbf{full} condition applies no projection. The \textbf{head-only} condition retains only the top-\(k\) singular directions. The \textbf{tail-only} condition removes them and retains only the residual component.

In implementation, we compute an equivalent economy decomposition inside the LoRA parameterization using QR factorization followed by SVD on the resulting \(r \times r\) matrix. This yields an exact projection of the current update into the chosen spectral subspace while keeping the LoRA parameter shapes fixed throughout training.

\begin{table}[t]
\centering
\caption{Family-level summary for training-time projection experiments  with \(k=16\).}
\label{tab:q4_projection_family_mean}
\begin{tabular}{llccc}
\hline
Run & Variant & ID mean & OOD mean & TRAP mean \\
\hline
Run-Direct & base      & 0.1388 & 0.0675 & 0.2938 \\
Run-Direct & full      & 0.2775 & 0.1150 & 0.5063 \\
Run-Direct & head-only  & 0.1625 & 0.0100 & 0.3050 \\
Run-Direct & tail-only & 0.1388 & 0.0675 & 0.2938 \\
\hline
Run-Deliberate & base      & 0.1388 & 0.0675 & 0.2938 \\
Run-Deliberate & full      & 0.3250 & 0.0663 & 0.5650 \\
Run-Deliberate & head-only  & 0.1975 & 0.0225 & 0.3175 \\
Run-Deliberate & tail-only & 0.1388 & 0.0675 & 0.2938 \\
\hline
\end{tabular}
\end{table}

\subsection{Supervision Corruption}
\label{app:shuffle_details}

To test whether stable preference learning depends on prompt--preference consistency, we corrupt a fraction \(\rho \in \{0.25,0.50,1.00\}\) of the training set using \texttt{shuffle\_pairs}. For each corrupted example, the original prompt is kept fixed, but its preference pair \((y^+,y^-)\) is replaced by the pair from another randomly selected training example. The pair is replaced as a whole; chosen and rejected responses are not shuffled independently.

Corruption is applied only to the training set. All evaluation splits remain clean. In Figure~\ref{fig:q5_shuffle_curves}, the point at \(\rho=0\) corresponds to the uncorrupted full-training result.

\begin{table}[t]
\centering
\caption{Family-level summary for supervision-corruption experiments using training-time \texttt{shuffle\_pairs}.}
\label{tab:q5_shuffle_family_mean}
\begin{tabular}{llccc}
\hline
Run & Corruption ratio \(\rho\) & ID mean & OOD mean & TRAP mean \\
\hline
Base & 0.0 & 0.1388 & 0.0675 & 0.2938 \\
\hline
Run-Direct & 0.25 & 0.1613 & 0.0550 & 0.4238 \\
Run-Direct & 0.50 & 0.0850 & 0.0075 & 0.2300 \\
Run-Direct & 1.00 & 0.0950 & 0.0363 & 0.2475 \\
\hline
Run-Deliberate & 0.25 & 0.2425 & 0.0525 & 0.3863 \\
Run-Deliberate & 0.50 & 0.2100 & 0.0663 & 0.4463 \\
Run-Deliberate & 1.00 & 0.0313 & 0.0200 & 0.2238 \\
\hline
\end{tabular}
\end{table}

The supervision-corruption experiment is an upstream behavioral probe. It does not, by itself, provide a complete spectral account of how corruption changes the strength of the spectral head, the thickness of the residual tail, or the trajectory of effective rank.

\end{document}